\documentclass[3p,times]{elsarticle}

\usepackage{amssymb}
\usepackage{amsfonts}
\usepackage{amsmath}
\usepackage{graphicx}
\usepackage{hyperref}       % hyperlinks
\hypersetup{colorlinks,linkcolor={blue},citecolor={blue},urlcolor={blue}}
\usepackage{url}
\usepackage[symbol]{footmisc}

\usepackage[figuresright]{rotating}

\usepackage[utf8]{inputenc}
\usepackage{fancyhdr}
\usepackage[titletoc,title]{appendix}
\usepackage[noend]{algpseudocode}
\usepackage{listings}
\usepackage{enumerate}
\usepackage{color}

\usepackage{array}
\usepackage{multirow}

\usepackage{caption}
\usepackage{subcaption}
\usepackage{pdflscape}

\definecolor{dkgreen}{rgb}{0,0.6,0}
\definecolor{gray}{rgb}{0.5,0.5,0.5}
\definecolor{mauve}{rgb}{0.58,0,0.82}

\begin{document}

\begin{frontmatter}




\title{A Framework for End-to-End Learning on Semantic Tree-structured Data}


\author{William Woof and Ke Chen\footnote[1]{Corresponding author}}

\address{Department of Computer Science, The University of Manchester, Manchester M13 9PL, U.K.\\
Email: \{William.Woof, Ke.Chen\}@manchester.ac.uk}

\begin{abstract}
%TODO: Update abstract
While learning models are typically studied for inputs in the form of a fixed dimensional feature vector, real world data is rarely found in this form. %, and appear in their natural form that effectively encodes a variety of intrinsic information.
In order to meet the basic requirement of traditional learning models, structural data generally have to be converted into fix-length vectors in a handcrafted manner, which is tedious and may even incur information loss.
A common form of structured data is what we term ``semantic tree-structures", corresponding to data where rich semantic information is encoded in a compositional manner, such as those expressed in \emph{JavaScript Object Notation }(JSON) and \emph{eXtensible Markup Language} (XML).
For tree-structured data, several learning models have been studied to allow for working directly on raw tree-structure data, however such learning models are limited to either a specific tree-topology or a specific tree-structured data format, e.g., synthetic parse trees.
In this paper, we propose a novel framework for end-to-end learning on generic semantic tree-structured data of arbitrary topologies and heterogeneous data types, such as data expressed in JSON, XML and so on.
Motivated by the works in recursive and recurrent neural networks, we develop exemplar neural implementations of our framework for the JSON format.
We evaluate our approach on several UCI benchmark datasets, including ablation and data-efficiency studies, and on a toy reinforcement learning task.
Experimental results suggest that our framework yields comparable performance to use of standard models with dedicated feature-vectors in general, and even exceeds baseline performance in cases where compositional nature of the data is particularly important. The source code for a JSON-based implementation of our framework along with experiments can be downloaded at \url{https://github.com/EndingCredits/json2vec}.
\end{abstract}


\end{frontmatter}

% \linenumbers

\section{Introduction}
\label{sect:intro}

In their natural form, real world data typically appears in a manner that effectively encodes semantic and structural information surrounding the underlying data. However, most traditional machine learning algorithms are only applicable to fixed-dimensional feature vectors, which requires handcrafted feature extraction from structured data such as trees, sets and graphs. Recently, novel deep learning models have been emerging to handle structured data towards avoiding handcrafted feature engineering, e.g., set networks \cite{zaheer2017setnet} and graph neural networks \cite{zhou2018graphnet}. Such learning models have been manifested in enhancing the capacity of learning models via exploitation of structural information and better performance in dealing with complex structured data.

%However,
Another ubiquitous form of data is what we refer to as ``semantic tree-structured data".
Semantic tree-structured data can be seen as a variant on more conventional tree structures, but where additional semantic information is incorporated into the structure.
In general, branch nodes are used to accommodate high-level objects, instantiated from various semantic classes, while leaf nodes represent a variety of primitives in different data formats. Importantly, the data contained within semantic tree-structures is typically significantly heterogeneous, both at a technical level and at a semantic level.
Although some tree-structured learning models have been studied, all of those so far are limited to a specific tree-topology, e.g., all the leaf nodes located in a layer of the same depth to the root \cite{HAMMER20041061,zhang2018t2v}, or a specific tree-structured data format, e.g.,
\emph{syntactic parse trees} widely used in nature language processing \cite{socher2013treebank,irsoy2013bidirectional,paulus2014belief,irsoy2014deeprecursive,tai2015improved,kokkinos2017tree}, where all objects at the same tree depth are of the same form.
To the best of our knowledge, there are no learning models that can directly deal with generic semantic tree-structured data of arbitrary topologies, where nodes in the same layer may contain entities of heterogeneous data types, although the importance of this emerging topic was highlighted (including example application for JSON data) and some theoretical analysis regarding universal approximation was studied in the recent work of \cite{pevny2019universal}.

In this paper, we present a learning framework, \emph{semantic tree-structured recursive learning architecture} (STRLA), for end-to-end learning\footnote{Although we focus on supervised learning in this paper, it is directly applicable to reinforcement learning as demonstrated in Section \ref{sub:rl-task}}. STRLA allows for inputting generic semantic tree-structured data, e.g. JSON data, directly and fulfils automatic feature extraction and model learning simultaneously.
Unlike the prior work of \cite{pevny2019universal}, our framework details how to construct a learning model for any semantic tree-structured format.
Additionally, our framework performs architectural construction at the prediction stage, allowing application to arbitrary data topologies without modification.
Motivated by the ideas underlying the deep set network \cite{zaheer2017setnet} and recurrent neural networks \cite{hochreiter1997long, tai2015improved}, we carry out our STRLA framework with two different neural implementations for the JSON data format, leading to different deep recursive neural networks of heterogeneous components.
In our experiments, we demonstrate the empirical strength of our neural implementation of STRLA based on several UCI benchmark datasets \cite{Dua:2019}, where we emulate JSON descriptions of data examples.
Our work presented in this paper provides an enabling technology which automatically creates a neural architecture for dealing with generic semantic tree-structured data, and alleviate the requirement to transmute such data to a fixed set of feature vectors: a missing technical component in all the ongoing AutoML paradigms \cite{AutoML}.

The main contributions in this paper are summarised as follows:
\begin{itemize}
\item We formalise the notion of a semantic tree-structure, which encompasses various hierarchical semantically-annotated data formats such as JSON and XML.
\item A novel yet generic framework, STRLA, is proposed for end-to-end supervised learning on arbitrary semantic tree-structured data, which simultaneously extracts structural and semantic features underlying  tree-structured data and model learning.
\item Exemplar neural implementations of our STRLA framework are developed for JSON and XML.% to be used as enabling techniques.
\item A thorough comparative study is conducted to demonstrate the effectiveness of our STRLA framework on a variety of data expressed in JSON.% and XML.
\end{itemize}

The rest of this paper is organised as follows. Section 2 reviews related works, and Section 3 describes a notational scheme that covers all the descriptive notations or languages of generic semantic tree-structured data. Sections 4 and 5 presents our STRLA framework and its different neural implementations, respectively. Section 6 describes experimental settings and reports the experimental results. Section 7 discusses the issues arising from our work, and the last section draws conclusions.

\section{Related work}
\label{sect:related}

%In general, recursive learning models related to tree-structured data may be divided into three categories in terms of various tasks: semantic parsing and syntax transformation, exploitation of syntactic structures and representation learning.

In this section, we make a connection to the previous works closely relevant to our work presented in this paper and highlight the main differences between those and ours.

Dated back to 1990's, recursive neural networks \cite{coller1996task-dependent,frasconi1998general} had been proposed to tackle the problems arising from structured data. Since then, those ideas have inspired the development of learning models for different structured data, e.g., \cite{zaheer2017setnet,zhou2018graphnet,socher2013treebank,irsoy2013bidirectional,paulus2014belief,irsoy2014deeprecursive,tai2015improved,kokkinos2017tree}.
Regarding tree-structured data, such ideas are widely utilised in learning models for semantic parsing and syntax transformation, e.g., \cite{socher2011parsing,rabinovich2017abstract,chen2018t2t}, and exploitation of syntactic structures, e.g, \cite{socher2013treebank,irsoy2013bidirectional,paulus2014belief,irsoy2014deeprecursive,tai2015improved,kokkinos2017tree,li2015tree}. In a broad sense, our work presented in this paper is also inspired by the general ideas of original recursive neural networks \cite{coller1996task-dependent,frasconi1998general} to deal with the compositional information encoded in semantic tree-structured data recursively.

For semantic parsing and syntax transformation, several recursive learning modes have been proposed, e.g., \cite{socher2011parsing,rabinovich2017abstract,chen2018t2t}. Such models make use of recursive properties to discover a tree-structured representation from input data such as 2-D images and sequential text/code, e.g., \cite{socher2011parsing,rabinovich2017abstract} or transform tree-structure data of one type into another, e.g., \cite{chen2018t2t}. Although such models share general ideas with ours in a broad sense in dealing with tree-structural information, those models work especially for semantic parsing and syntax transformation tasks, which are distinct from those problems tackled by our framework presented in this paper; i.e., our framework is proposed for end-to-end supervised learning directly from raw semantic tree-structured data rather than generation of a tree-structured representation from input data of other forms.

So far, most of recursive learning models related to tree-structured data have been studied to exploit the tree-structured information underlying sequential text formed with syntactic rules for various \emph{natural language processing} (NLP) tasks, e.g., \cite{socher2013treebank,irsoy2013bidirectional,paulus2014belief,irsoy2014deeprecursive,tai2015improved,kokkinos2017tree,li2015tree}.
To this end, each sentence in text is first converted into a syntactic parse tree and then a learning model is developed to exploit additional structural/syntactic information underlying text for various NLP tasks ranging from sentiment analysis to question answering and relation classification. Those recursive learning models working on parse trees in an end-to-end manner have turned out to be very effective for various NLP tasks. However, a syntactic parse tree represents the syntactic structure of strings according to some context-free grammar, which is simply a class of specific semantic tree-structure data. Hence, such models cannot deal with generic semantic tree-structured data such as JSON, XML and HTML where intermediate nodes may accommodate objects/entities instantiated from different semantic classes and leaf nodes may represent various types of primitives in different data formats. Moreover, there are other challenging problems in generic semantic tree-structured data beyond syntactic parse trees, e.g., a path from a root to a specific node in generic semantic tree-structured data provide useful global contextual information for supervised learning and hence needs to be explored in a learning model. To the best of our knowledge, the global path information has yet to be explored in all the existing recursive learning models working on synthetic parse trees although the local parent-child relation information was used in some recursive learning model such as the global belief recursive neural Networks \cite{paulus2014belief}. Nevertheless, those ideas behind the existing recursive learning models for syntactic parse trees generally inspire our work presented in this paper. In particular, the neural implementation of our framework is motivated by the tree-structured LSTM networks \cite{tai2015improved}, a recursive learning model originally proposed for syntactic parse trees. Given the fact that as several existing recursive learning models, e.g.,
\cite{socher2013treebank,irsoy2013bidirectional,paulus2014belief,irsoy2014deeprecursive,tai2015improved,kokkinos2017tree,li2015tree}, have been developed especially for syntactic parse trees by making good use of their specific properties, our work presented in this paper does not target those problems arising from syntactic parse trees but tackles the challenging issues arising from generic semantic tree-structured data of arbitrary topologies and heterogenous data types for end-to-end supervised learning.

There are yet other methods that deal with tree-structured data for representation learning, e.g., recursive self-organizing networks \cite{HAMMER20041061} and Tree2vector \cite{zhang2018t2v}. Unlike those aforementioned recursive learning models for tree-structure discovery/transformaton and supervised NLP tasks, these unsupervised learning models tend to deal with tree-structured data via decomposition of the structures into their basic constituents for feature extraction \cite{HAMMER20041061}  or learn a vectorial representation of a fixed length for tree-structured data \cite{zhang2018t2v}. However, such methods assume consistent semantic meaning between the nodes at any given level of a tree and the same depth of all the leaf nodes to the root. Hence, such methods can only work on specific tree-structured data of an ad hoc topology and homogeneous data types but cannot deal with generic semantic tree-structured data. Unlike our framework working in an end-to-end fashion for supervised learning, moreover, such learning models cannot be applied to supervised learning directly. For supervised learning on tree-structured data, such models yield a vectorial representation of tree-structure data only and other supervised learning models have to be employed based on the vectorial representation. In other words, two stages, feature extraction and model learning, have to be undergone, which is suboptimal for supervised learning, apart from its limited applicability to specific semantic tree-structured data stated above.

During the process of this manuscript, an unpublished work mentioned in \cite{pevny2019universal} emerged, which appears closely relevant to ours presented in this paper.
%With the same goal as ours, the work demonstrates how to deal with JSON data directly for end-to-end classification.
Based on their code and our personal communication with the authors (as the technical details of their approach are not included in \cite{pevny2019universal}), we understand that this work reflects a preliminary effort in handling semantic tree-structured data in the JSON format towards automatic machine learning, focusing the inherent problem of multi-tier variadicy of JSON and other semantic tree structure forms.
However, as implemented, their approach can only cope with the semantic tree-structured data of a fixed data schema, meanwhile our recursive architecture is established via dynamic construction for semantic tree-structured data of arbitrary topologies.
Moreover, while \cite{pevny2019universal} provides a specific technical implementation for JSON data, we provide a more general framework, and introduce a number of additional concepts such as element paths.

\section{Semantic Tree Structure Description}
\label{sect:stree}

To understand how our framework can be applied to arbitrary semantic tree-structured data, we first describe a meta-notational scheme of the generic semantic tree-structure with the syntax of the \emph{Backus Naur form} (BNF) \cite{Knuth:1964:BNF}. Specific descriptive notations or mark-up languages of semantic tree structures, e.g., JSON and XML, can be instantiated or derived from this meta-notational scheme. We then give an illustrative example of semantic tree-structured data expressed in JSON to facilitate our presentation in the next section. Finally, we highlight the element path, an important concept associated with structural and compositional information.

\subsection{Meta-notational Scheme}
\label{subsect:notation}

In general, we specify generic semantic tree-structured data as serialisation of data that can be represented with a variety of different hierarchical semantically-annotated data formats that conforms to several requirements as follows:
\begin{itemize}
    \item The data can be decomposed into individual data elements, each of which can be uniquely identified as belonging to one of a finite number of pre-specified types.
    \item Elements are allowed to contain other elements within them recursively. Moreover, there is at least one of the identified types consisting of a list of (potentially wrapped) elements, with no additional data. Such types are named ``containers" and the remaining types are referred as to ``primitives".
    %Ideally the primitive types should consist of simple data such as strings, and numbers.
    \item There is a wrapper format used to denote a given element with a name (given as a string). Optionally, this wrapper may also denote an element with a description.
\end{itemize}

\begin{figure*}[th]
    \centering
\begin{lstlisting}
<STS> ::= <primitive> | <container>

<primitive> ::= <p1> | <p2> | ...
; <primitive>s are the raw data of the data structure

<container> ::= <c1> | <c2> | ...
<c1> ::= <open-c1> <t-c1> *(<sep-c1> <t-c1>) <close-c1>
<c2> ::= <open-c2> <t-c2> *(<sep-c2> <t-c2>) <close-c2>
; and so on for <c3>, <c4>, ...
; Each container <c1>, <c2>, ... is a list of elements of types from the set <t-c1>, <t-c2>, ...
;  where each <t-ci> is equal to one of "<STS>", "<wrapped-STS>", or "<STS> | <wrapped-STS>"

<wrapped-STS> ::= fn[ <name>, (<description>), <STS> ]
; The wrapper must contain a <name> and a single <STS>
; -- optionally the wrapper can contain some additional <description>
\end{lstlisting}
    \caption{Meta-notational descriptive scheme for semantic tree structures with the BNF syntax. ``\texttt{fn[a, b, c]}" specifies a combination of the elements \texttt{a}, \texttt{b}, and \texttt{c}. ``\texttt{*(d)}" denotes repeat of an identical element \texttt{d} multiple times up to the changes denoted by the ordinal numberings.}
    \label{fig:sts_spec}
\end{figure*}

\begin{figure*}[th]
    \centering
    \begin{subfigure}[t]{0.4\textwidth}
        \centering
        \includegraphics[width=\textwidth]{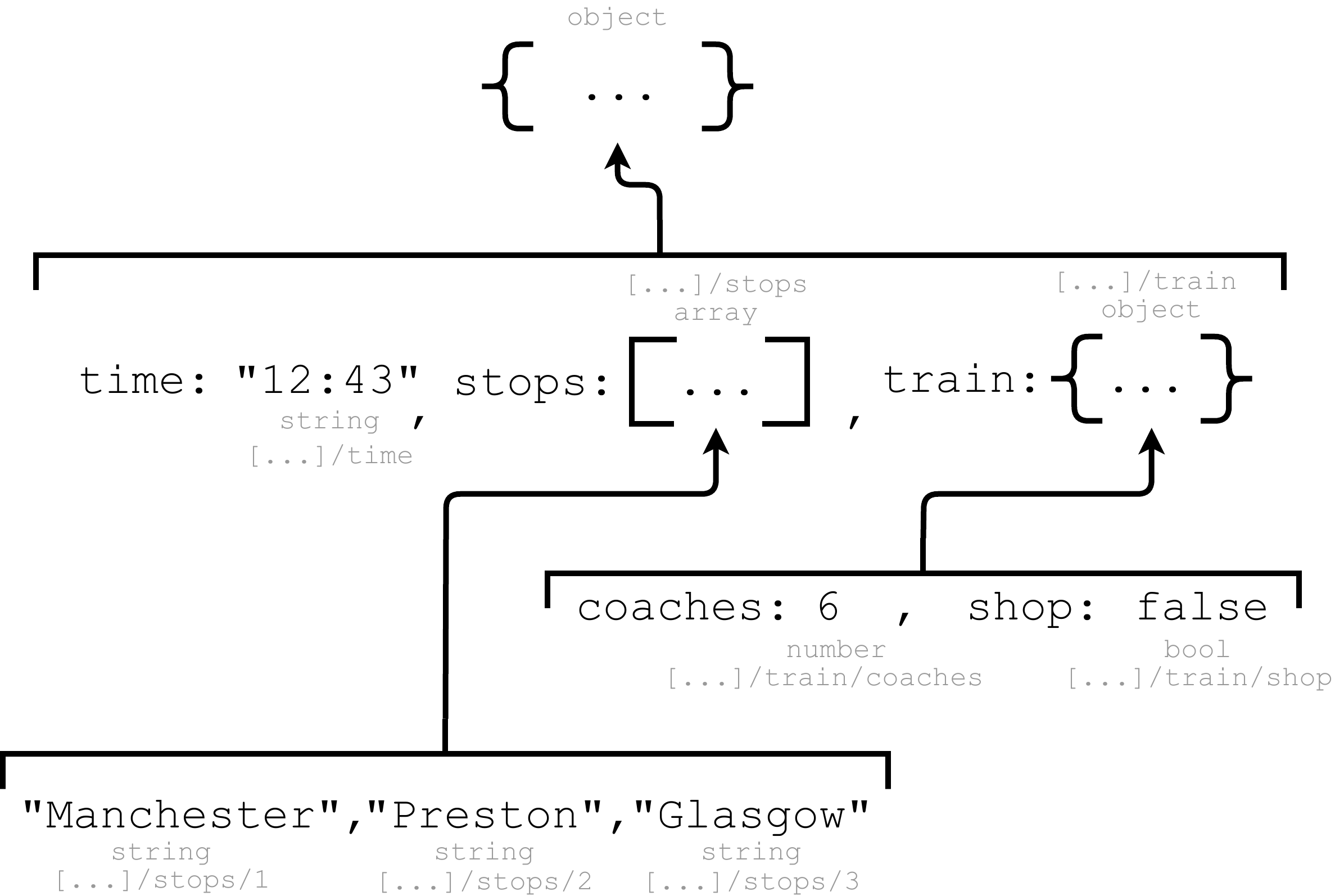}
        \caption{}
    \end{subfigure}
    \begin{subfigure}[t]{0.4\textwidth}
        \centering
        \includegraphics[width=\textwidth]{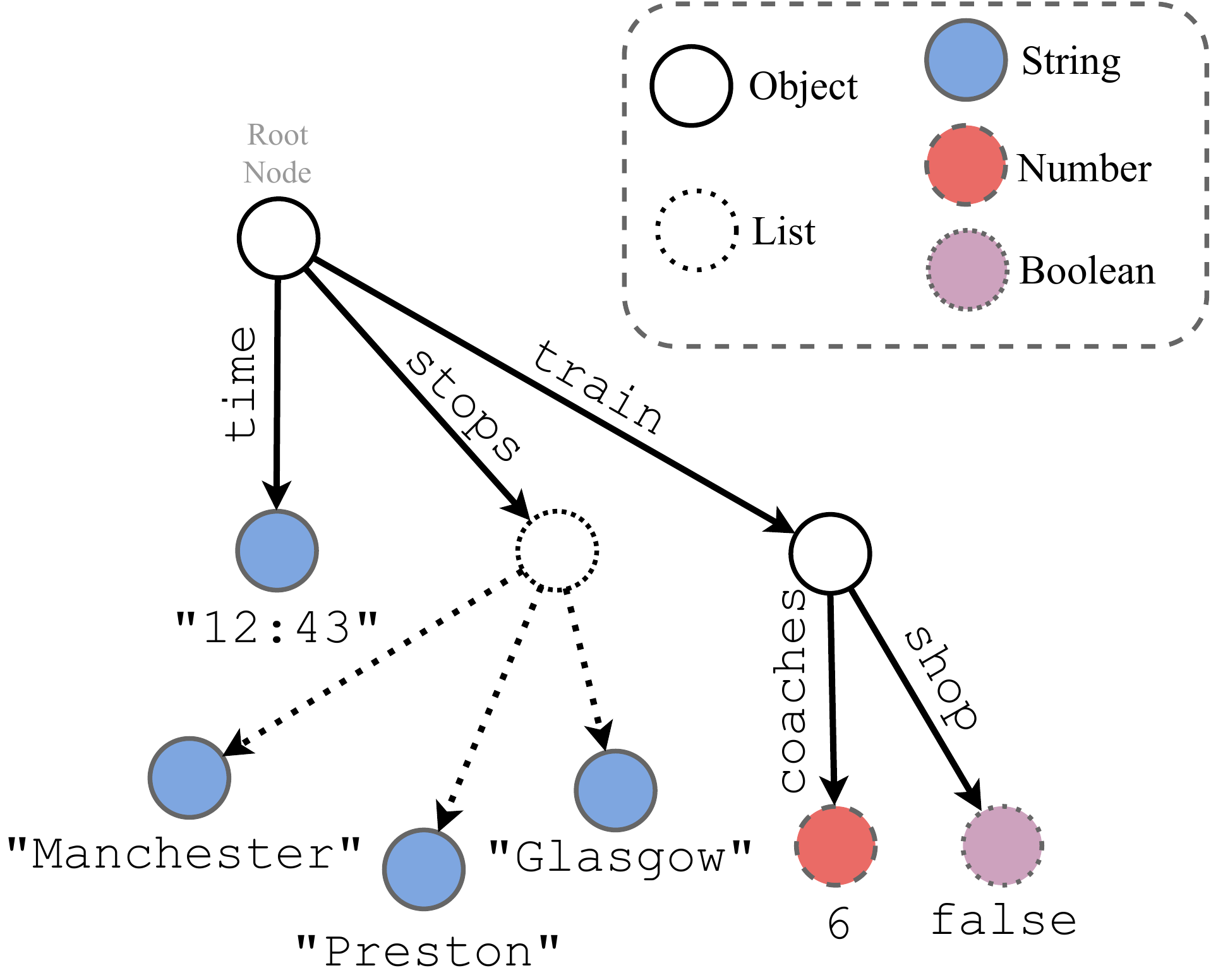}
        \caption{}
    \end{subfigure}
    \caption{The ``train journey" represented (a) in hierachical form (including element paths) and (b) as a tree structure.}
    \label{fig:train-journey}
\end{figure*}

To describe the semantic tree structures formally, we present a meta-notational scheme with the BNF syntax, as depicted in Figure~\ref{fig:sts_spec}.
As demonstrated in Figure~\ref{fig:train-journey}, data in such a format can be viewed as a semantic tree where the root node indicates the whole object. The branch and leaf nodes are used to accommodate the container and primitive elements respectively, and edges between nodes are annotated with semantic tags.
Unlike a traditional tree, however, only leaf nodes in a semantic tree are associated with any data items, while branch nodes serves purely for hierarchical organization reflecting compositional relationship between parent and sibling nodes, with the exception of added information in the form of certain additional ``description" tags within certain formats. Furthermore, the semantic tree generally has a heterogeneous nature; i.e. both branch and leaf nodes may accommodate objects/data belonging to different classes and various data types.

Furthermore, the meta-notational scheme entails a number of core concepts such as element types, primitives, containers and wrapped-elements. This scheme enables us to approach the problem in a implementation-agnostic fashion so as to build a general framework from which we can derive architectures for a variety of different human-readable data formats of semantic tree structures. For instance, the specification of JSON format can be achieved by instantiating the corresponding components in the meta-notational scheme with its constituents; i.e., three primitive elements, \texttt{<number>}, \texttt{<string>} and \texttt{<boolean>}, and two container elements, \texttt{<object>} and \texttt{<array>}. Thus, the JSON format can be specified formally with this meta-notational scheme. Likewise, XML and other semantic tree structure formats can be specified in the same manner. To facilitate the reader's understanding, we illustrate the specification of JSON and XML formats derived from the meta-notational scheme in Appendix \ref{appA}.

%\begin{figure*}[th]
%    \centering
%    \includegraphics[width=0.4\textwidth]{train-journey.pdf}
%    \caption{An illustrative example of semantic tree-structured data, the ``train journey".}
%    \label{fig:train-journey}
%\end{figure*}

\subsection{Illustrative Example}
\label{subsect:example}

To understand the semantic tree-structured data, we employ a ``Train Journey" scenario as an illustrative example regarding the specification of this semantic class and an object instantiated from this class.

The ``Train Journey" scenario can be described with several pieces of information organised in different categories, \emph{departure-time}, \emph{stops}, and \emph{train-info: number-of-carriages, is-there-a-shop}. It can be viewed as a semantic class organised with a semantic tree where separate data elements ``tagged" with different names via wrappers. For a concrete train journey, an object can be instantiated from the class. For instance, given a concrete train journey from Manchester to Glasgow in U.K.,
the time the train leaves the first station is \emph{12:43}, it stops at \emph{Manchester}, \emph{Preston}, and \emph{Glasgow}, and the information regarding this train is that it has \emph{6} carriages and \emph{no} shop on board. Hence, this specific train journey object can be expressed by a semantic tree shown in Figure \ref{fig:train-journey}.

Represented in the JSON format (c.f. Figure \ref{fig:json}), the above example illustrates how the different features enable the creation of rich data structures. The elements of types \emph{string}, \emph{number} and \emph{boolean} contain the actual data in leaf nodes of the structure, and the two containers, \emph{array} and \emph{object}, organise the raw data elements in branch nodes to create a hierarchy. The range of different element types allows the structure to  represent individual data snippets more precisely.
Finally, the tagged data elements enable the annotation of elements within the hierarchy with semantic tags. Hence, the ``train journey" object can be represented in the JSON format as follows:
\begin{verbatim}
{
    "time": "12:43",
    "stops": [ "Manchester", "Preston", "Glasgow" ],
    "train": { "carriages": 6, "shop": false }
}
\end{verbatim}

\subsection{Element Path}
\label{subsect:path}

As shown in Figure \ref{fig:sts_spec}, wrappers are used to annotate sub-structures via a name field, effectively specifying the elements contained within. However, the immediate name of an element does not always fully describe its semantic context. For example, if a given sub-structure is labelled as an ``address", it is not clear whether the data corresponds to the address of an individual, a building, an organisation etc. Nevertheless, the ambiguity would be avoidable if it is known that this ``address" structure is part of a larger ``employee" structure. Hence, the context or the compositional information conveys very important structural information in semantic tree-structured data, which is referred as to \emph{element path} in this paper.

Formally, we define a path, $p_e$ of an element $e$, with respect to some parent element $e_0$, to be the sequence of names of wrappers by moving up the hierarchy along with the associated branches from $e$ to $e_0$.
Thus, an element path is formed by a sequence of strings denoted as a single string generated by the concatenation of those stings for names of wrappers  in reverse order separated by slashes. Assume that \texttt{c}, \texttt{b}, \texttt{a} are strings for names of wrappers for an element $e$ within a parent element $e_0$, its element path $p_e$ is represented by a single string, ``\texttt{a/b/c}".
The construction of this path follows the same rules as for existing concepts such as Xpath for XML \cite{clark1999xml}, and JSONPath \cite{goessner2007jsonpath} for JSON, which are typically used for addressing individual elements. However, we note that these element addresses are equally usefully for describing semantic context.

More specifically, for the ``train journey" object described in Section \ref{subsect:example}, the \emph{is-there-a-shop} element, the value \texttt{false}, is directly wrapped with a wrapper with name \texttt{shop}, which is contained within a JSON object. This JSON object is wrapped with another wrapper, now with name \emph{train}. Therefore, the element path of the \emph{is-there-a-shop} within the root is ``\texttt{train/shop}".
Nevertheless, the \emph{departure-time} element, with value \texttt("12:43") is wrapped with only a single wrapper, with name \texttt{time}, hence its element path within the root is simply ``\texttt{time}".
Similarly, despite being contained in a further list container, the \emph{stops} elements, \texttt{Manchester}, \texttt{Preston}, and \texttt{Glasgow}, are only within a single wrapper, and hence all have the element path ``\texttt{stops}".

In practice, it can often be worthwhile to add extra information to this element path, such as container types or ordinal positions in lists.
Under this scheme, the element paths for \texttt{Manchester}, \texttt{Preston}, and \texttt{Glasgow} would become \texttt{stops/1}, \texttt{stops/2}, and \texttt{stops/3}, providing a bit more context for these elements.
Since the mapping of wrappers can be achieved by mapping from one definition of semantic tree structures to another before any application of our framework\footnote{Specifically, this can be done by assigning `dummy' wrappers to elements. For example, we could choose to map any lists of (unwrapped) elements \texttt{[ a, b, ... ]} to lists of elements wrapped with their ordinal position in the list \texttt{[ 1:a, 2:b, ... ]}.}, we treat this an implementation detail.
In our presentation, we hence use only the strict definition of element path defined above.

Here, we emphasise that the proper use of element paths in our framework presented in the next section is a salient characteristic that significantly distinguishes ours from all the related works for tree structures reviewed in Section~\ref{sect:related}.

\section{Framework}
\label{sect:framework}

In this section, we present the \emph{semantic tree-structure recursive learning algorithm} (STRLA), which is a generic framework for end-to-end supervised learning directly from semantic tree-structured data (as it was specified in Section \ref{sect:stree}).
%
%STRLA effectively combines network construction with feed-forward
%
This framework stands in a format-agnostic manner and hence serves as a general solution for addressing the common issues underlying any semantic tree-structures at a functional level.

As a semantic tree-structured document is intrinsically expressed as tree structure, our framework takes the general idea behind the recursive neural networks \cite{coller1996task-dependent,frasconi1998general}, a well-known architecture for tree-structural learning, and adapts it to address the common issues in learning from semantic tree-structured data.
A recursive neural network employs a common neural-network component for each node of the tree by assigning each one a latent ``hidden-state" vector based on the data associated with the current node and the hidden-states of its child nodes (where they exist).
This neural network component is applied recursively to each node of a tree, moving up from the leaves of the tree to the root, such that every node would be eventually assigned a hidden state.
In order to develop an effective recursive learning architecture to tackle our problem, we consider those salient features and common issues underlying generic semantic tree-structure data. As a result, we identify several aspects of semantic tree-structure data that must be considered, but which differ from other types of tree-structured data (such as syntactic parse trees for natural language).
%that have been handled with various recursive neural networks for different learning tasks as reviewed in Section \ref{sect:related}.

These aspects specific to semantic tree-structured data are summarised as follows:
\begin{enumerate}
    \item The data associated with leaf nodes may be heterogeneous, conforming to a variety of different primitive types specified by the specific data format.
    \item The branch nodes of a semantic tree may also be heterogeneous, corresponding to different container types.
    \item The number of children of each branch node is not fixed, which allows a branch node to have any number of children.% corresponding to different container or primitive types.
    \item Additional structural or contextual information encoded in the element path as defined in Section \ref{subsect:path} is carried with each node, which we would exploit to facilitate the learning from semantic tree-structured data.
\end{enumerate}

%TODO: rework
To address the aforementioned issues arising from semantic tree-structured data, we come up with a general recursive learning framework that takes into account all the requirements:
1. To address the issue on heterogeneity of data, our framework would employ a set of functions, \{$f_p$\}, to deal with different primitive types separately.
2. Similarly, for branch nodes, we employ a set of functions, \{$f_c$\}, to be applicable to all of the permissible container types.
3. Since the number of children of each node is not fixed, branch/container node functions are required to deal with the input of an arbitrary number of latent vectors.
4. As highlighted previously, we would also exploit the contextual information carried in the element path. To this end, we incorporate the element path information into each of all the aforementioned functions, \{$f_p$\} and  \{$f_c$\}, where path information is supplied as an additional input.

%TODO hence we need a set of functions
As described above, our framework is reliant on a set of functions, \{$f_p$\} and \{$f_c$\}, known as the element embedding functions, which map from each element to a latent hidden-state vector.
For each primitive type, a function is employed so as to yield a fixed-length latent representation that encodes all the information relevant to any node of that type, along with a path description.
Formally, for each primitive type $p$ a primitive embedding function is required to generate a $m$-dimensional latent representation for leaf nodes as follows:
\begin{equation}
\label{eq:primitive}
f_p: {\mathcal X}_p \times \mathcal P  \rightarrow \mathbb R^m,
\end{equation}
where ${\mathcal X}_p$ is the space corresponding to valid forms of element data of type $p$, and $\mathcal P$ is the space of all the valid element paths.
The number of functions, \{$f_p$\}, required is the same as that of all the permissible primitive types, where each of different primitive type is dealt with by a separate function.
Similarly, for each container type $c$, an embedding function to yield a $m$-dimensional latent representation for branch nodes is required.
Unlike those functions made for leaf nodes, a function for branch nodes has to yield a latent representation upon the input of both those latent representations (hidden-state vectors) of its immediate children, and the local description information associated with the element (if there is any), e.g., attributes of a tag in XML.
Formally, each $f_c$ must be of the form
\begin{equation}
\label{eq:container}
f_c: (\mathbb R^m)^{\mathbb N} \times \mathcal P \times \mathcal D \rightarrow \mathbb R^m,
\end{equation}
where $(\mathbb R^m)^{\mathbb N}$ denotes the space of sets of $m$-dimensional latent representations, $\mathcal P$ is the space of all the valid element paths, and $\mathcal D$ is the space of all the valid description on the local information of branch nodes, which is purely dependent on the specific descriptive notation or mark-up language in question.

\begin{figure}
     \centering
     \begin{subfigure}[b]{0.3\textwidth}
         \centering
         \includegraphics[width=\textwidth]{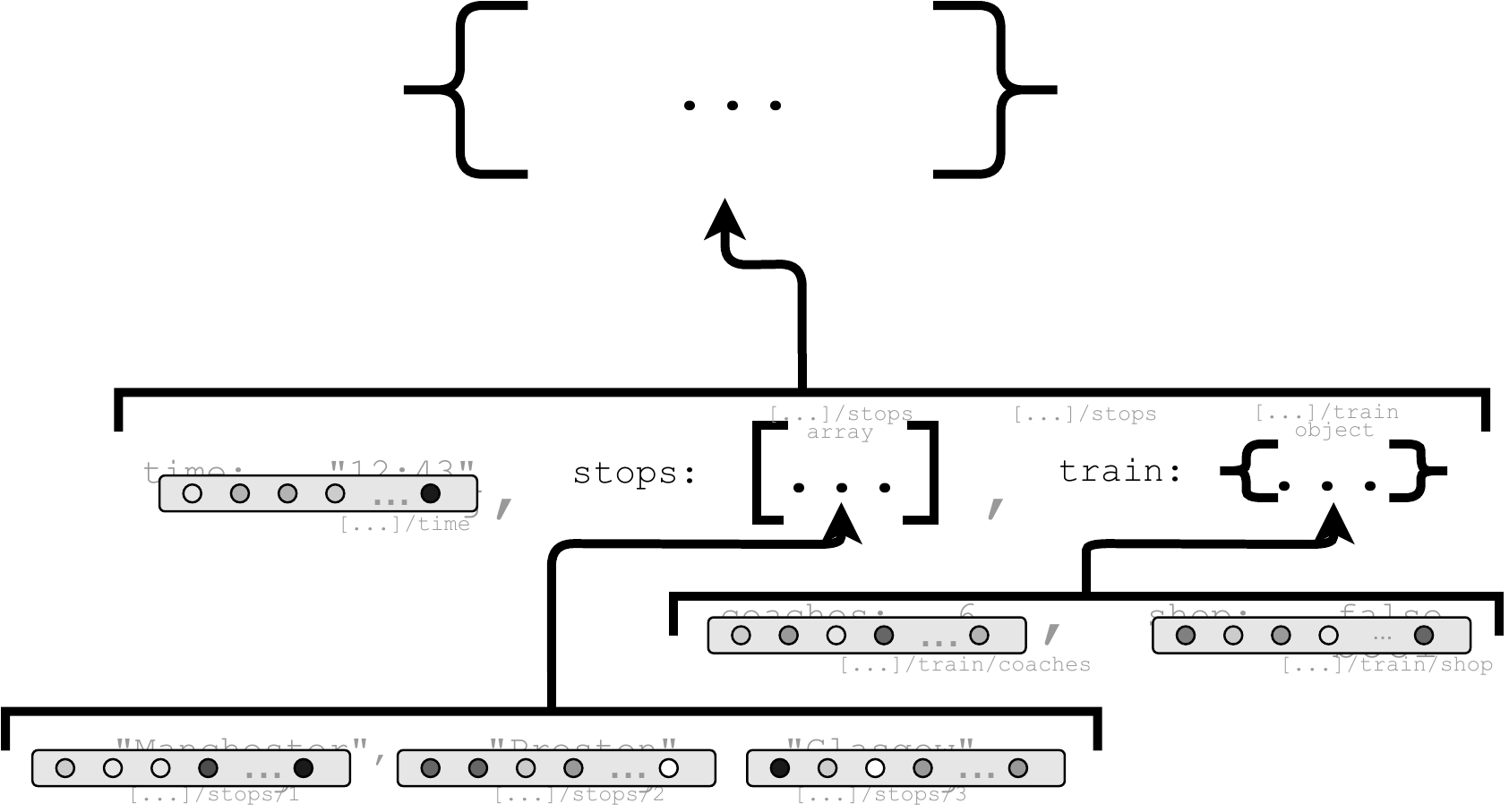}
         \caption{}
         \label{subfig:json_iterative_process_prim}
     \end{subfigure}
     \hfill
     \begin{subfigure}[b]{0.3\textwidth}
         \centering
         \includegraphics[width=\textwidth]{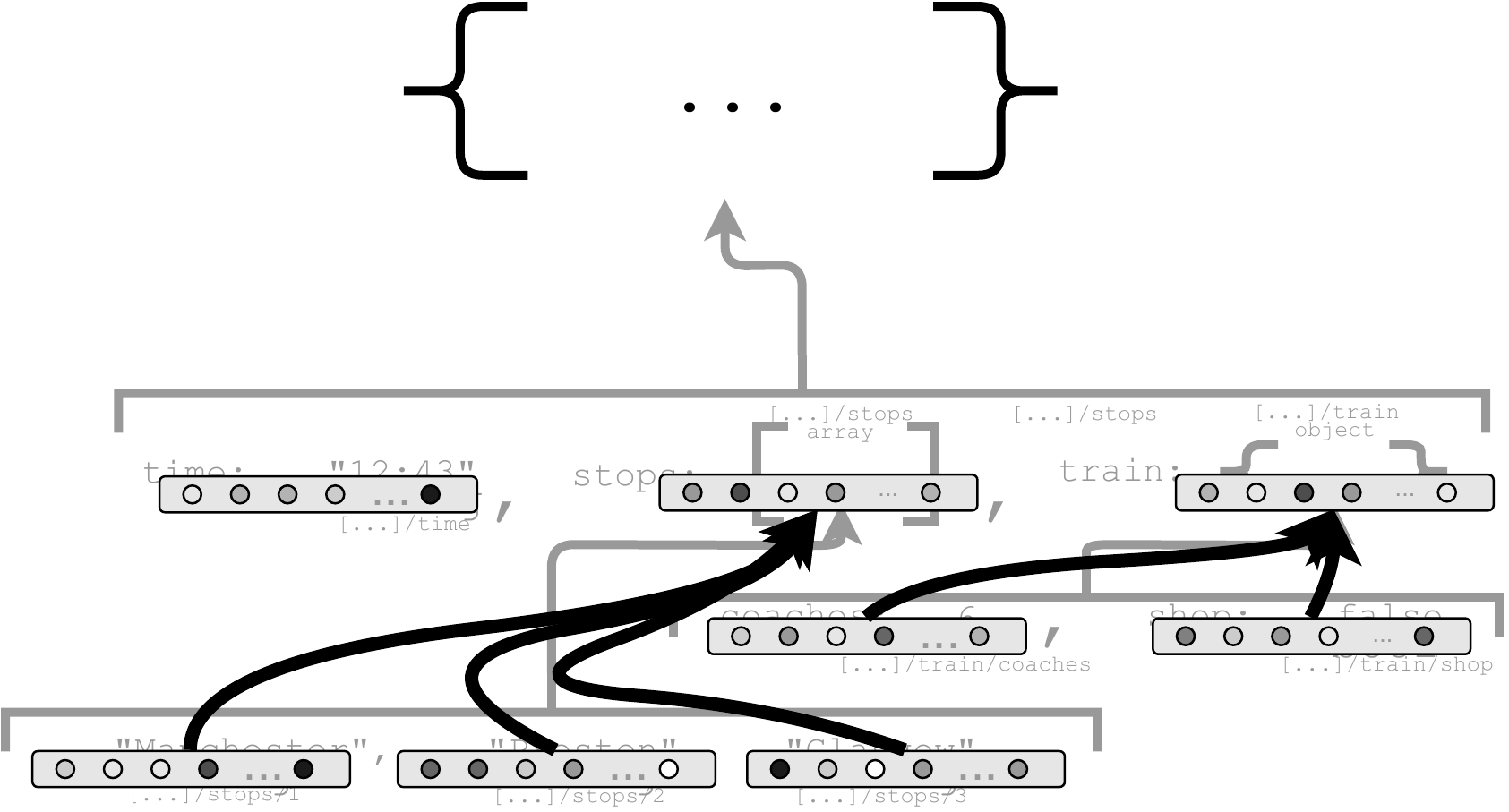}
         \caption{}
         \label{subfig:json_iterative_process_cont}
     \end{subfigure}
     \hfill
     \begin{subfigure}[b]{0.3\textwidth}
         \centering
         \includegraphics[width=\textwidth]{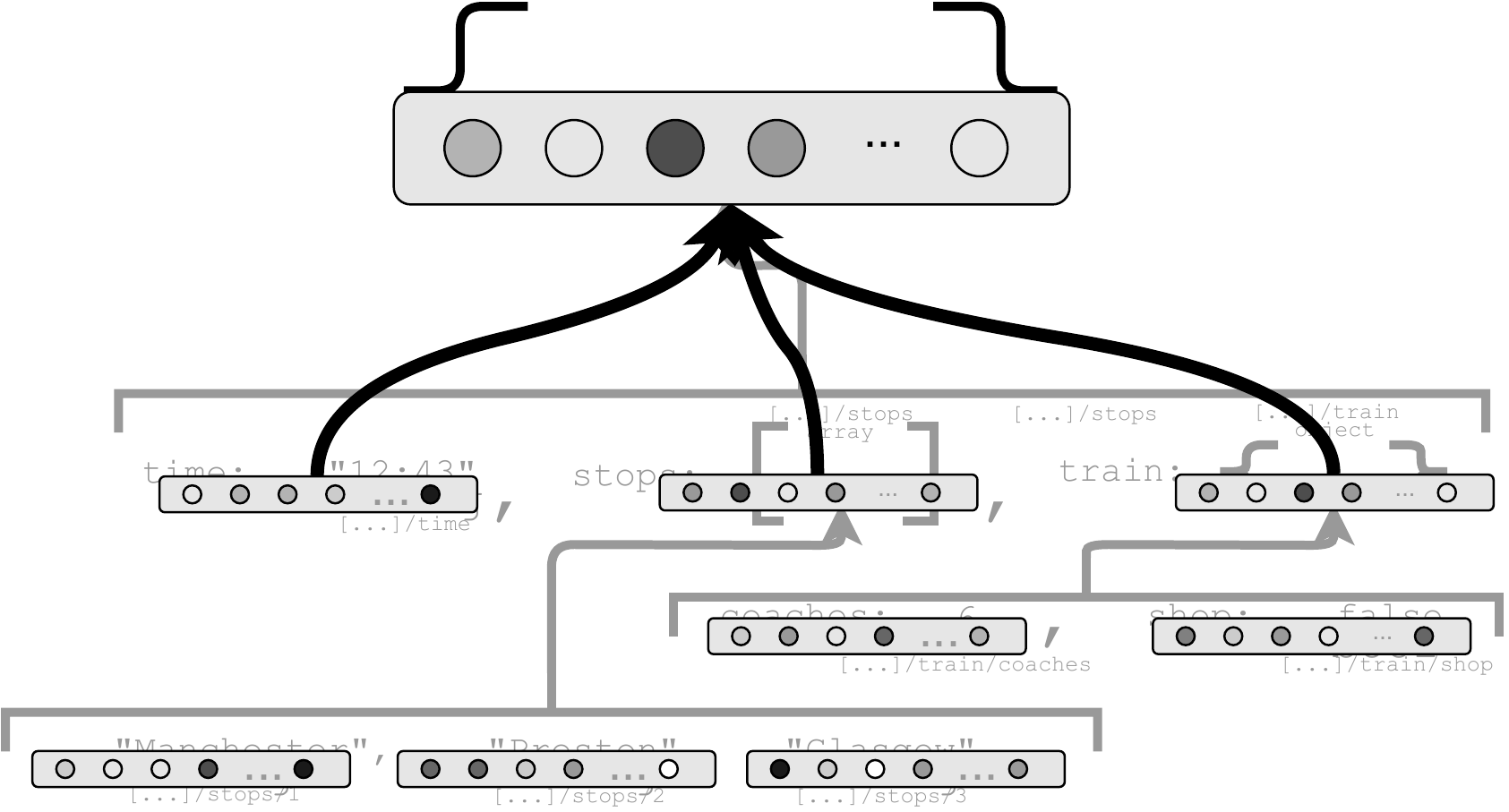}
         \caption{}
         \label{subfig:json_iterative_process_iter}
     \end{subfigure}
        \caption{Iterative application of embedding functions applied to JSON example }
        \label{fig:json_iterative_process}
\end{figure}

Given a semantic tree structure ${\pmb x}_{st}$, rooted at element $e_0$, then an $m$-dimensional latent representation for ${\pmb x}_{st}$ is obtained by iteratively applying those functions \{$f_p$\} and \{$f_c$\} to the nodes of the tree, as shown in Figure~\ref{fig:json_iterative_process}:
The latent representation of each leaf node is given by applying $f_{t_e}$ to the node data, $e$, and node path, $p_e$, via $f_{t_e}(e, p_e)$, where $t_e$ is the type of $e$ (Figure~\ref{subfig:json_iterative_process_prim}).
For branch nodes $e$ whose child elements $c_1,...,c_{n_e}$ have all been assigned a latent hidden-state, $\pmb h_{c_1},... \pmb h_{c_{n_e}}$, the latent representation is given by applying $f_{t_e}$ to the set of child hidden-states, along with node path, $p_e$, and description, $d_e$, via $f_{t_e}( \{ \pmb h_{c_1},... \pmb h_{c_{n_e}} \}, p_e, d_e)$ (Figure~\ref{subfig:json_iterative_process_cont}).
This latter step is applied repeatedly as more branch nodes are mapped, until finally the root node, $e_0$ is mapped to a latent hidden state vector (Figure~\ref{subfig:json_iterative_process_iter} ).

%Each leaf node corresponds to a primitive data element $e$ with type $t_e$ and path $p_e$. The latent representation $\pmb h_e$ of $e$ is given by $f_{t_e}(e, p_e)$ (Figure~\ref{subfig:json_iterative_process_prim}).
%Each branch node corresponds to a container data element $e$ with child elements, $c_1,...,c_{n_e}$, description $d_e$, type $t_e$ and path $p_e$. If the children of $e$, $c_1,...,c_{n_e}$ have been assigned a latent representation, then the latent representation of $e$ is given by $f_{t_e}( \{ \pmb h_{c_1},... \pmb h_{c_{n_e}} \}, p_e, d_e)$  (Figure~\ref{subfig:json_iterative_process_cont}).
%The previous step is repeated until the root element $e_0$ has been assigned a latent representation (Figure~\ref{subfig:json_iterative_process_iter} ).

\begin{figure}
     \centering
     \begin{subfigure}[b]{0.45\textwidth}
         \centering
         \includegraphics[width=\textwidth]{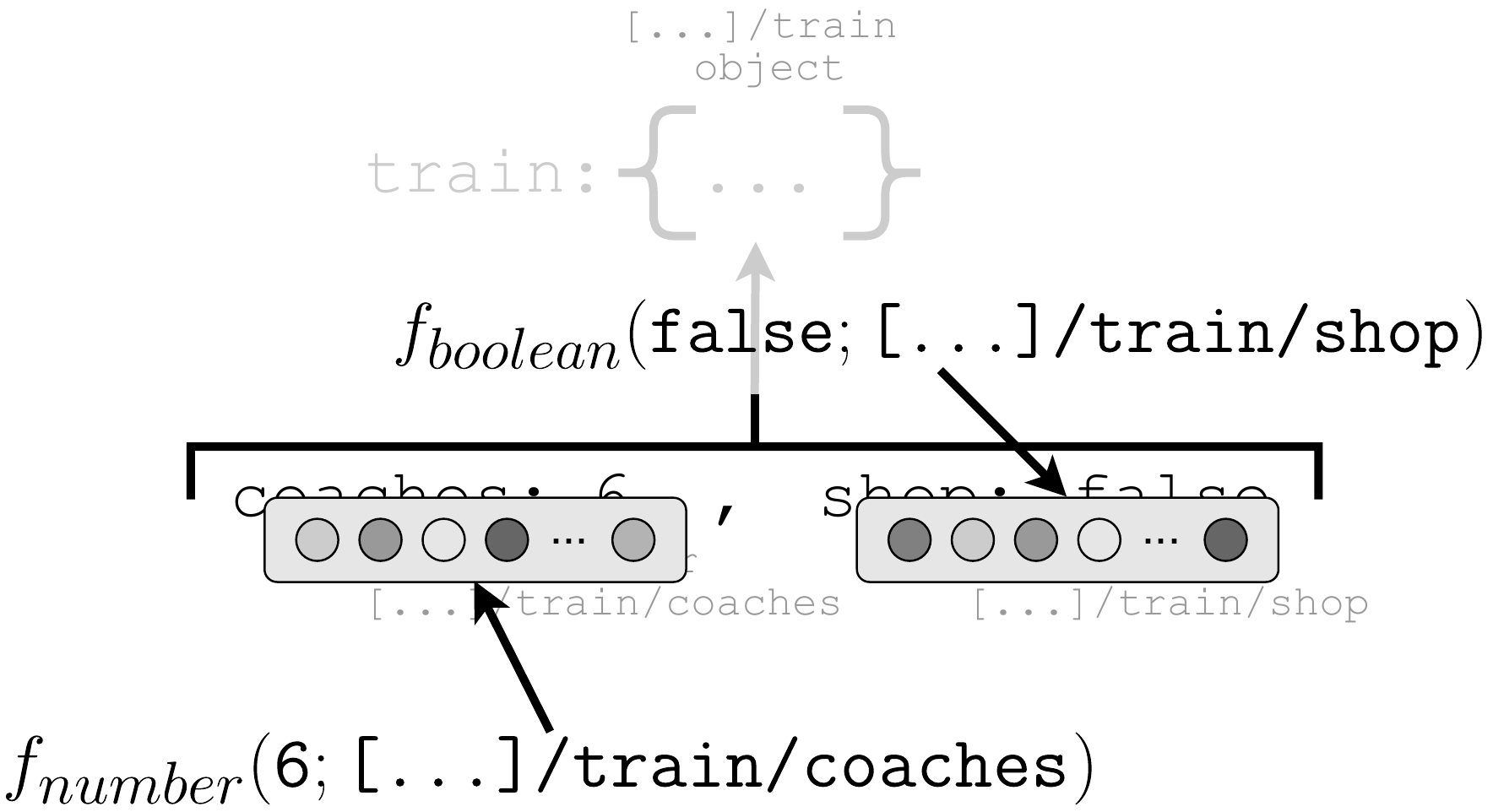}
         \caption{}
     \end{subfigure}
     \begin{subfigure}[b]{0.35\textwidth}
         \centering
         \includegraphics[width=\textwidth]{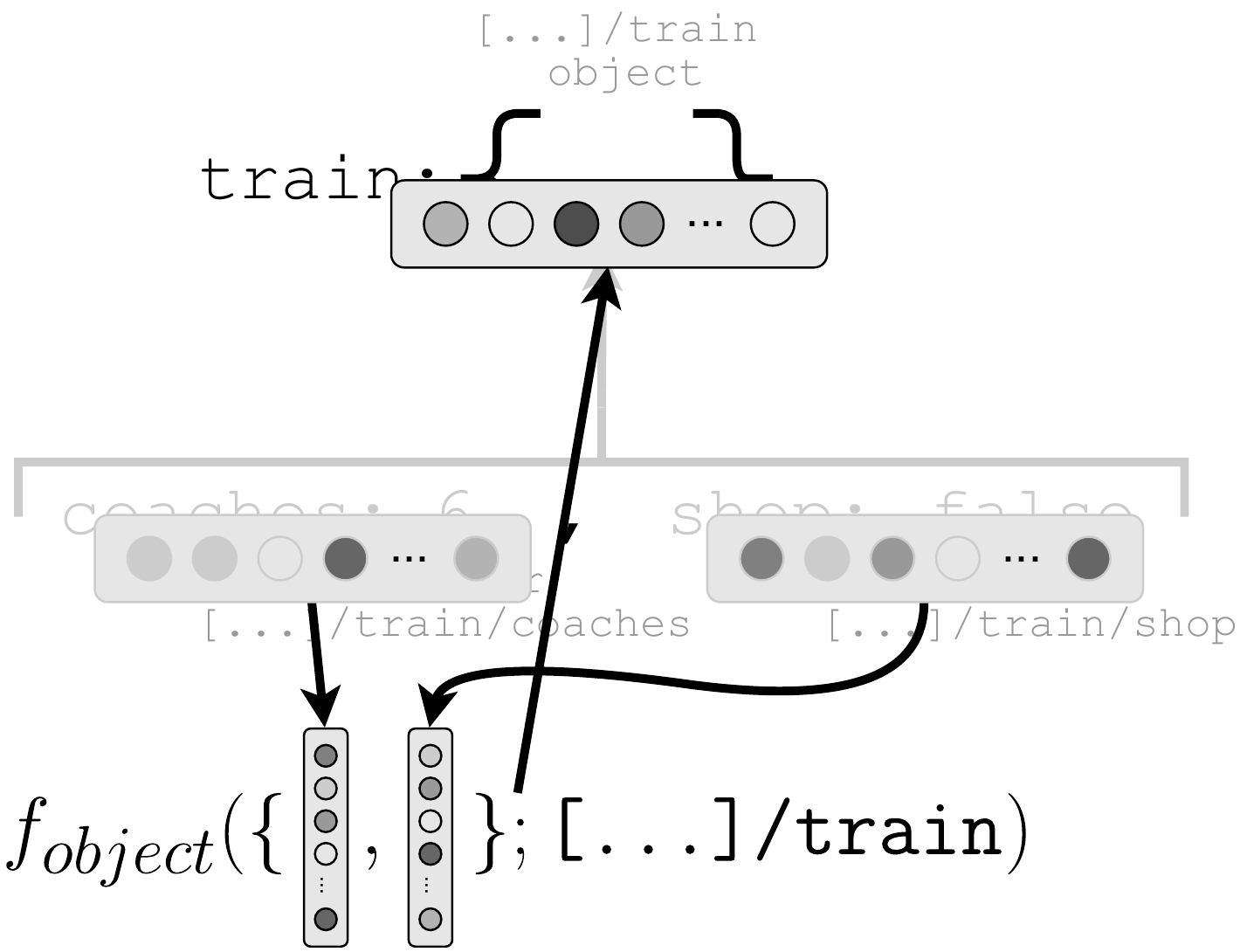}
         \caption{}
     \end{subfigure}
        \caption{Application of embedding functions to a) primitive (leaf) elements, and b) container (branch) elements.  }
        \label{fig:json_embedding}
\end{figure}

This process can be described via a recursive function $R(\circ, \circ, \circ)$, which maps from a given element, $e$, along with its path $p_e$, and description $d_e$, to an $m$-dimensional latent representation:
Given an element $e$, then the type $t_e$ of $e$ can be determined by the formatting rules of the specific descriptive notation or mark-up language.
If $e$ is a leaf node belonging to a primitive type, $p$, then the primitive embedding function, $f_p(\circ, \circ)$, is directly applied to $e$, along with $p_e$ to obtain its latent representation,
$$\pmb h_e = f_t(e, p_e)$$
Otherwise, if $e$ is a branch node belonging to a container type, $c$, then $e$ contains a number of child elements, $c_1, ..., c_{n_e}$.
For a given child element $c_i$, if $c_i$ is a wrapped element with name $n_i$ and description $d_{c_i}$, then the path $p_{c_i}$ is set to $p \texttt{/} n_i$. Otherwise if $c_i$ is unwrapped then the child path, $p_{c_i}$, is set to $p_e$ and $d_{c_i}$ is set to $\emptyset$.
The recursive function $R$ is applied to each child $c_i$ along with the child path and description, $p_{c_i}$ and $d_{c_i}$, to obtain a set of child latent representations: $\mathbf h_{1},\cdots,\mathbf h_{n_e}$.
The container embedding function, $f_c(\circ, \circ, \circ)$, is applied to the set $\mathbf h_{1},\cdots,\mathbf h_{n_e}$, along with $p$ and $d$, to obtain its latent representation,
$$\pmb h_e = f_c \big ((\pmb h_{1},\cdots,\pmb h_{n_e}), p_e, d_e \big )$$
Applying $R$ to the root element $e_0$ of a given semantic tree structure via $R(e_0, \emptyset, \emptyset)$ yields a latent vector, $\pmb h_{e_0}$, for the root of the tree, which can be taken as the latent representation of the tree.

For supervised learning the embedding functions, \{$f_p(\circ, \circ)$\} and \{$f_c(\circ, \circ, \circ)$\}, must be inferred from training data, $D=\{X_{st}, Y\}$.
%
%We construct a full parametric model $F({\pmb x}_{st}, \Theta_F)$, via $F({\pmb x}_{st}, \Theta_F)$
%.
This is fulfilled by applying a final parametric mapping function, $f: \mathbb R^m \rightarrow Y$, to the latent state associated with the root of the tree.
The full model is given by the function $F(x_{st} ~|~ \Theta_R, \Theta_f) := f \big ( R(e_0, \emptyset, \emptyset ~|~ \Theta_R )~|~\Theta_f \big )$, where $\Theta_R$ is a collective notation of all the parameters in those parametric functions of \{$f_p(\circ, \circ)$\} and  \{$f_c(\circ, \circ, \circ)$\}.
Thus, the end-to-end supervised learning from semantic tree-structured data is done by minimising a proper loss function, $ l \bigg (F\big (X_{st} ~|~ \Theta_R, \Theta_f \big), Y \bigg )$, with respect to the parameters $\Theta_R$ and $\Theta_f$. Typically, the mean squared error loss is used for regression, and the 0/1 loss is used for classification. 

For reinforcement learning, as same as done in supervised learning, our STRLA framework can be directly employed to learn approximating its value or Q function working on semantic tree-structured data in an end-to-end manner, which will be demonstrated in Section \ref{sub:rl-task}.

If deep neural networks are employed as parametric models, the optimisation of a loss function can be done via gradient-based optimisation, using back-propagation through structure \cite{coller1996task-dependent}.
Moreover, it has been proven in \cite{pevny2019universal} that the learned $F(\circ)$ function operating based on appropriately chosen \{$f_p(\circ, \circ)$\} and  \{$f_c(\circ, \circ, \circ)$\} functions acts as a universal approximator for semantic tree-structured data. Thus, our proposed framework is capable of acting as a general recursive learning architecture for supervised learning from any semantic tree-structured data. On the other hand, our proposed framework is primarily at a functional level and there are several non-trivial issues in its implementation such as developing proper deep neural networks to meet the requirements of various embedding functions, \{$f_p(\circ, \circ)$\} and  \{$f_c(\circ, \circ, \circ)$\}, incorporating the element path information into those deep neural networks and dealing with specifics of a concrete notational format describing semantic tree-structured data.

%In summary, the integration of different embedding networks for a given notational format such as the JSON and the XML and a feed-forward neural network with the latent representation of the root (i.e., its hidden state vector only in the LSTM based implementation) as its input leads to a learning architecture for supervised learning on any semantic tree-structured data expressed in the specific notational format. The learning is done in an end-to-end manner by optimising all the parameters together with a proper method, e.g., on-line gradient descent, on a training dataset (c.f. Sections \ref{sect:framework}).

\section{Neural Implementation}
\label{sect:implement}

While our framework presented in Section \ref{sect:framework} is generic and format-agnostic, an implementation of this framework has to take into account the specific aspects of the target notational format used for describing semantic tree-structured data. As described in Section \ref{sect:framework}, our framework for end-to-end supervised learning works via function $F(\circ)$ operating based on embedding functions, \{$f_p(\circ, \circ)$\} and \{$f_c(\circ, \circ, \circ)$\}, and output function $f(\circ)$. In our work, we employ a linear transformation for the final output transformation $f(\circ | \Theta_f)$. Thus, the main challenges lie in how to implement those embedding functions, \{$f_p(\circ, \circ)$\} and  \{$f_c(\circ, \circ, \circ)$\}.
In this section, we come up with exemplar implementations for the commonly-used notational format, JSON. We first address a common issue on how to incorporate the element path information into an implementation of any notational formats. Then, we present the implementations for the JSON format, respectively, where the element path information is exploited.

\subsection{Encoding the Element Path}
\label{subsect:e-path}

As described in Section \ref{subsect:path}, the element path provides the certain level of contextual information, which can be used in avoiding possible semantic ambiguities in semantic tree-structure data. In the following, we describe our implementations for incorporating this path information into the embedding functions, \{$f_p(\circ, p_e)$\} and \{$f_c(\circ, p_e, \circ)$\}.

%Explain in terms of attention function?
In our methods, we encode all the possible element paths in $\mathcal P$ via a dictionary that can be constructed based on training data. While $\mathcal P$ is infinite, the number of paths within the data will be finite.
For embedding functions \{$f_p(\circ, p_e)$\} and \{$f_c(\circ, p_e, \circ)$\}, we use the case-wise functions, where the functions regarding a given type $p$ (or $c$) and given path $p_e$ are as follows:
$$f_p(\circ, p_e) = f_{p, p_e}(\circ),$$
$$f_c(\circ, p_e, \circ) = f_{c, p_e}(\circ, \circ),$$
where each path $p_e$ is associated with a set of functions, \{$f_{p, p_e}(\circ)$\} and \{$f_{c, p_e}(\circ, \circ)$\}.
Now, we consider two different strategies to construct these functions.

\subsubsection{Path specific weights}
\label{subsubsect:path-specific-weights}

%...
We use parametric models of the same learning architectures for all paths, but having different learnable parameters decided by the element path, $p_e$:
$$f_p(\circ, p_e ) := \hat{f}_p(\circ ~|~ \Theta_{p, p_e})$$
$$f_c(\circ, p_e, \circ ) := \hat{f}_c(\circ, \circ ~|~ \Theta_{c, p_e})$$
where $\Theta_{p, p_e}$ and $\Theta_{c, p_e}$ are instances of parameters for the models $\hat{f}_p$ and $\hat{f}_c$ for given path $p_e \in \mathcal P$.
%
%The idea here is that a separate copy of model parameters are retained for each unique path. When the embedding function is applied, those parameters (and only those parameters) corresponding to the path of the node to be embedded are used.
%This is applied to each element separately,
%
In other words, we use separate network weights for each different path.

Encoding the element path information in this manner simplifies the implementation of the embedding functions specified in Eqs. (\ref{eq:primitive}) and (\ref{eq:container}) such that our neural implementation can deal with the semantic aspects directly without loss of structural information underlying semantic tree-structured data. It mimics the natural approach to constructing architectures combining multiple separate input sources of information, where separate network weights are used for different inputs, and is also effectively equivalent to the method employed in \cite{pevny2019universal}.
Hence, we focus on this method of encoding path information throughtout the majority of our experiments.
However, we emphasise that this is not the only method to encode the element path information, and in Section~\ref{subsubsect:path-specific-funcs} we discuss an extension of this idea. Additionally in \ref{sect:discuss} we discuss how variations on this method could be used to exploit other aspects of the data.

%There are alternative ways to encode the element path information as discussed in Section \ref{sect:discuss}, we emphasise that the manner described above is simple yet generic regardless of any notational formats.

\subsubsection{Path specific functions}
\label{subsubsect:path-specific-funcs}

One issue with path specific weights is that the same function is used for all elements of the same type. However, as well as variations between different types, even primitive data of the same type within a semantic tree structure may differ substantially. For example, in the ``Train Journey" example of Section~\ref{subsect:example}, the strings \texttt{"12:43"} and \texttt{"Manchester"} contain very different data, even though they are both of the \emph{string} type.

Here, we propose an extension to path specific weights, where instead of using the same base functions $\hat{f}_p$ and $\hat{f}_c$ for all paths, we employ a set of functions $\{ \hat{f}_{p,1}, ..., \hat{f}_{p, n_p} \}$ and $\{ \hat{f}_{c,1}, ..., \hat{f}_{c, n_c} \}$, which we apply on a path-specific basis.
To do this, for each type $t$ we employ a mapping $\mathcal{D_t}: \mathcal P \rightarrow \{ \hat{f}_{t,i} \}_{i=1...n_t}$ mapping from the space of paths, $\mathcal P$, to a set of functions $\{ \hat{f}_{t,1}, ..., \hat{f}_{t, n_t} \}$. We then use the function corresponding to the mapping applied to the given element path:
$$f_p(\circ, p_e ) := \hat{f}_{p, \mathcal D_p(p_e)}(\circ, p_e ~|~ \Theta_{p, p_e})$$
$$f_c(\circ, p_e, \circ ) := \hat{f}_{c, \mathcal D_c(p_e)}(\circ, p_e, \circ ~|~ \Theta_{c, p_e})$$
The mapping functions $\mathcal D_t$ are given via a user-defined mapping dictionary, which maps from each path to a function template.
For ease of application, in our implementation, we employ a psuedo-JsonPath \cite{goessner2007jsonpath} format, which allows wildcards to be used to specify multiple paths. %Examples are shown in

This can be considered an extension of the path-specific weights of Section~\ref{subsubsect:path-specific-weights}, where the topology of the underlying network is allowed to vary for different parts of the data schema. However, a downside to this method is that it relies on a manually-specified user-defined mapping from paths to function templates.

\subsection{JSON Recursive Networks}
\label{subsect:json-net}

As illustrated in Figure \ref{fig:json}, when expressed as a semantic tree-structure, the JSON format consists of five types: two container types,``array'' and ``object'', and three primitive types, ``number'', ``string'' and ``boolean''. As there are different requirements in implementing those primitive and container embedding functions, we employ proper neural networks to implement those embedding functions of different types. Unless otherwise stated, we use the path-specific weights of Section~\ref{subsubsect:path-specific-weights} for incorporating paths.

\subsubsection{Primitive Embedding Networks}
\label{subsect:json-primitive}

For all three primitive types, the embedding task is simply building a mapping from input of ``\emph{number}'', ``\emph{string}'' and ``\emph{boolean}'' to an $m$-dimensional latent representation.

For the \emph{number} type, we use a linear neural network to implement it. Let
${\rm lin}(z,\theta)$ denote the linear neural network:
${\rm lin}(z,\theta)= {\pmb w}z + \pmb b$, where $\theta= \{\pmb w, \pmb b\}$, $z$ is the scalar input, and $\pmb w$  and $\pmb b$ are $m$-dimensional weight and bias vectors.
Given a leaf-node $e$ containing a scalar number, $x$, and its element path, $p_e \in \mathcal P$,  the embedding of the number type is carried out by
\begin{equation}
\label{eq:json-num}
  f_{num}(x|\Theta_{num,p_e}) = {\rm lin} \big ( \hat{x}, \Theta_{num,p_e} \big ),
\end{equation}
where $\hat{x}=\frac{x-\mu_p}{\sigma_p}$ ($\mu_p$ and $\sigma_p$ are the mean and the standard deviation of all $x$) if there are more than one number-type leaf nodes in training data of path $p_e$, and $\hat{x}=x$ otherwise.

For the \emph{boolean} type, its value, $x$, can be converted into integers:  ${\rm int}(x) = 1$ if $x=``\texttt{true}"$ and ${\rm int}(x) = 0$ if $x=``\texttt{false}"$. Thus, we would use the same linear neural network formulated in Eq. (\ref{eq:json-num}) to carry it out; i.e., for a leaf-node $e$ containing a boolean value, $x$, and its element path, $p_e \in \mathcal P$,
\begin{equation}
\label{eq:json-bool}
  f_{bool}(x|\Theta_{bool,p_e}) = {\rm lin} \big ( {\rm int}(x), \Theta_{bool,p_e} \big ).
\end{equation}

For the $string$ type, the embedding task is more complex as a string consists of numerous characters and its length may vary. To tackle this sequence modeling problem, we employ a simple single-layer recurrent neural network of LSTM units \cite{hochreiter1997long} to carry out the string embedding task at the character level since LSTM has turned out to be one of the most effective techniques in dealing with ordered sequences of variable lengths. For a leaf-node $e$ carrying a string, $s=\{s_1,\cdots,|s|\}$, and having the element path, $p_e \in \mathcal P$, the embedding function is carried out by the LSTM network of $m$ units as follows:
%$s=\{s_1,s_2,\cdots,s_e \}$
\begin{equation}
\label{eq:json-str}
  f_{str}(s|\Theta_{str,p_e}) =
    \frac{1}{|s|}\sum_{i=1}^{|s|} {\pmb h}^{\rm LSTM}_i,
\end{equation}
where $\Theta_{str,p_e}$ is the collective notation of all the learnable parameters in the LSTM network (see Appendix \ref{sec:LSTM_eqns} for details) and ${\pmb h}^{\rm LSTM}_i$ is the hidden state vector when character $i$ is sequentially input to the LSTM network, which is obtained by
$$
%\bigg ({\pmb c}^{\rm LSTM}_i, {\pmb h}^{\rm LSTM}_i \bigg )
{\pmb h}^{\rm LSTM}_i
 \Leftarrow {\rm LSTM} \bigg (\bigg \{{\rm lin}(s_1,\Theta_{s_1}),\cdots,{\rm lin}(s_i,\Theta_{s_i}) \bigg \}, \Theta_{str,p_e} \bigg),
$$
where ${\rm lin}(s_j,\Theta_{s_j})$ leads to an $m$-dimensional latent representation of character $j$ for $j=1,\cdots,i$.
%$$
%\bigg ({\pmb c}^{LSTM}_i, {\pmb h}^{LSTM}_i \bigg )
%= {\rm LSTM} \bigg (\big \{{\rm lin}(s_1,\Theta_{s_1,p_e}),\cdots,{\rm lin}(s_i,\Theta_{s_i,p_e}) \big \}, \Theta_{str,p_e} \bigg),
%$$
%where ${\rm lin}(s_j,\Theta_{s_j,p_e})$ leads to an $m$-dimensional latent representation of character $j$ for $j=1,\cdots,i$.

\subsubsection{Container Embedding Networks}
\label{subsect:json-container}

For the containers, an \emph{array} is a list of raw JSON entries while an \emph{object} is a list of wrapped JSON entries in the form of name-JSON pairs. In the JSON format, there is no additional description for branch nodes, i.e. $\mathcal D = \emptyset$. Therefore, we need to implement $f_c(\circ|\Theta_{c, p_e})$ instead of $f_c(\circ, \circ|\Theta_{c, p_e})$. As containers are associated with branch nodes that may have a variable  number of children in a semantic tree, an implementation of container embedding functions has to tackle this issue. Here, we present two different implementations based on deep set networks \cite{zaheer2017setnet} and different LSTM learning architectures \cite{hochreiter1997long,tai2015improved}.

\paragraph{Deep-set based Implementation}
Deep sets \cite{zaheer2017setnet} has been proposed to tackle the learning problem when the input is in the form of permutation invariant sets, where the number of elements in a set may vary. If we treat two container types, array and object, as sets, the deep set networks thus become an enabling technique to carry out two container embedding functions in the same manner. A deep set network consists of two components: \emph{element-embedding} and \emph{pooling} nets. In our work, the element-embedding net is implemented by a fully-connected feed-forward neural network and the pooling net is carried out with the averaging sum of the output of all the element-embedding nets applied to different elements in a set. Let ${\rm Lin}({\pmb z}, \theta)$ denote a linear hidden layer: ${\rm Lin}({\pmb z}, \theta)= W{\pmb z} + {\pmb b}$, where $\theta= \{W, \pmb b\}$, $\pmb z$ is its input, and $W$  and $\pmb b$ are its weight matrix and bias vector, and
${\rm ReLu}({\pmb x}) = \max \{{\pmb x}, {\pmb 0} \}$ denote the output of a hidden layer consisting of ReLU units based on the corresponding linear hidden layer, i.e., ${\pmb x}={\rm Lin}({\pmb z}, \theta)$. Thus, the output of any hidden layers in the element-embedding net can be achieved recursively; i.e., if there are $L$ hidden layers, their output are
$$
{\pmb h}^{(l)} = {\rm ReLu} \bigg ( {\rm Lin} \big ({\pmb h}^{(l-1)}, \theta^{(l-1)} \big ) \bigg ),~~
l = 1, \cdots, L
$$
where ${\pmb h}^{(0)}$ is the external input to the element-embedding net.
Assume that a branch node, $e$, of a container type, $c$, has $n_e$ children whose latent representations are ${\pmb h}_{1},\cdots, {\pmb h}_{n_e}$.  The output of the element-embedding net with the parameters, $\Theta_{c}=\big \{ \theta_c^{(1)}, \cdots, \theta_c^{(L-1)} \big \}$,  corresponding to a child is ${\pmb h}_n^{(L)}$ for $n=1,\cdots,n_e$. That is, all the element-embedding nets for a container type, $c =$ ``\emph{array}" or ``\emph{object}", share the same parameters, $\Theta_{c}$, for invariant permutation.
Accordingly, the pooling net is implemented by
$$
\phi_{ave}({\pmb h}_1,\cdots, {\pmb h}_{n_e}|\theta_{ave,c})
={\rm ReLu} \bigg (
 {\rm Lin} \bigg (\frac 1 {n_e} \sum_{n=1}^{n_e}{\pmb h}_n^{(L)}, \theta_{ave,c} \bigg ) \bigg ).
$$
By incorporating the element path, $p_e \in \mathcal P$, two container embedding functions are carried out based on the deep set networks described above in the same manner; i.e., for $c =$ ``\emph{array}" or ``\emph{object}",
\begin{equation}
\label{eq:json-container}
  f_{c}\big ({\pmb h}_1,\cdots, {\pmb h}_{n_e}|\Theta_{c,p_e} \big ) = {\rm Lin} \bigg (\phi_{ave}\big ({\pmb h}_1,\cdots, {\pmb h}_{n_e}|\theta_{ave,c} \big ), \theta_{c,p_e} \bigg ),
\end{equation}
where $\Theta_{c,p_e}$ is a collective notation of all the parameters; i.e., $\Theta_{c,p_e}= \big \{ \Theta_c, \theta_{ave,c},  \theta_{c,p_e} \big  \}$.

\paragraph{LSTM based Implementation}

A potential drawback of the deep-set based implementation is that two container types are treated the same. However, an array is different from an object;
an object contains wrapped/named elements and is hence permutation/order invariant, while an array consists of raw elements whose order may be an important information source. Therefore, we employ the vanilla LSTM  \cite{hochreiter1997long} to implement the array embedding function and the \emph{Child-Sum Tree-LSTM} (SumLSTM) to carry out the object embedding function. SumLSTM is actually one of Tree-LSTM models \cite{tai2015improved}, a variant of the LSTM, which can recursively deal with data expressed in syntactic parse trees for different learning tasks. %Details of the LSTM and the SumLSTM are described in Appendix \ref{appB}.
Given a branch node of the array type, $e$, having the element path, $p_e \in \mathcal P$, and $n_e$ children, the \emph{array} embedding function is carried out with the LSTM as follows:
\begin{equation}
  f_{array} \bigg ( \big \{({\pmb c}_1, {\pmb h}_1), \cdots, ({\pmb c}_{n_e}, {\pmb h}_{n_e}) \big \}| \Theta_{array,p_e} \bigg ) =
     {\pmb h}^{\rm LSTM}_{n_e},
    %\bigg ( {\pmb c}^{\rm LSTM}_{n_e}, {\pmb h}^{\rm LSTM}_{n_e} \bigg ),
\end{equation}
where $\Theta_{array,p_e}$ is the collective notation of all the parameters in the vanilla LSTM for an element path, $p_e \in \mathcal P$.
%${\pmb c}^{\rm LSTM}_{n_e}$ and
${\pmb h}^{\rm LSTM}_{n_e}$ is achieved recurrently via
\begin{equation*}
  %\bigg ({\pmb c}^{\rm LSTM}_k, {\pmb h}^{\rm LSTM}_k \bigg )
  {\pmb h}^{\rm LSTM}_k
 \Leftarrow {\rm LSTM} \bigg (\big \{{\pmb c}_1||{\pmb h}_1, \cdots , {\pmb c}_k||{\pmb h}_k \big \}, \Theta_{LSTM} \bigg ),
\end{equation*}
where ``$||$" is the concatenation operator of two vectors and $\big ( {\pmb c}_k, {\pmb h}_k \big )$ are the memory cell and hidden state vectors of the SumLSTM unit corresponding to child $k$ of the branch node, $e$ (see Appendix \ref{sec:SumLSTM_eqns} for details). For a branch node of the object type, $e$, having the element path, $p_e \in \mathcal P$, and $n_e$ children, the \emph{object} embedding function is carried out with the SumLSTM as follows:
\begin{equation}
f_{obj} \bigg ( \big \{({\pmb c}_1, {\pmb h}_1), \cdots, ({\pmb c}_{n_e}, {\pmb h}_{n_e}) \big \}| \Theta_{obj,p_e} \bigg ) =
    \bigg ( {\pmb c}^{\rm SumLSTM}_e, {\rm Lin} \big ( {\pmb h}^{\rm SumLSTM}_e, \theta_{obj,p_e} \big ) \bigg ).
\end{equation}
Here, $\Theta_{obj,p_e}= \{\Theta_{SumLSTM}, \theta_{obj,p_e}\}$  is the collective notation of all the parameters in the parametric model implementing the object embedding function, including all the parameters in the SumLSTM, $\Theta_{SumLSTM}$, shared by all branch node of the object type and the parameters of linear transformation, $\theta_{obj,p_e}$, merely applied to the hidden state vector, ${\pmb h}^{\rm SumLSTM}$, in order to incorporate the element path information into the object embedding function. As described in Appendix \ref{sec:SumLSTM_eqns},
${\pmb c}^{\rm SumLSTM}_e$ and ${\pmb h}^{\rm SumLSTM}_e$ are achieved via
\begin{equation*}
  \bigg ({\pmb c}^{\rm SumLSTM}_e, {\pmb h}^{\rm SumLSTM}_e \bigg )
 \Leftarrow {\rm SumLSTM} \bigg (\big \{({\pmb c}_1, {\pmb h}_1), \cdots , ({\pmb c}_{n_e}, {\pmb h}_{n_e} ) \big \}, \Theta_{SumLSTM} \bigg ).
\end{equation*}
It is worth stating that when any child of a branch node is a leaf node (primitive), its corresponding ``memory cell vector", $\pmb c$, is always set to zero while the latent embedding representation of this primitive node (c.f. Section \ref{subsect:json-primitive}) is treated its corresponding``hidden state vector", $\pmb h$, in the above SumLSTM implementation of the object container.

Here, we emphasise that the implementations for the JSON format presented above is easily extensible to any other notational formats specified in Figure \ref{fig:sts_spec}, as demonstrated in Appendix~\ref{ap:xml-net} for the XML format.

\section{Experiments}
\label{sect:experiment}

In this section, we describe our experimental settings on a comparative study and report the experimental results on several benchmark datasets. 

\subsection{Experiments on UCI datasets}

To demonstrate our proposed models, we compare the two JSON models described above to several baselines on a number of different classification tasks. In particular, we choose a number of well-known benchmark datasets from the UCI machine learning repository \cite{dua2019uci}.
%
%The University of California Irvine Machine Learning Repository  is a collection of machine learning datasets, intended for use within research into machine learning systems.
%The UCI repository contains a large number of distinct datasets making it a good source for testing new (and existing) machine learning algorithms over a wide range of different domains.
%
Since these datasets appear in ``tabular" form, we synthesize their corresponding JSON versions by deriving JSON structures from the intrinsic semantic-relationship between different attributes for each dataset.

%All models feature a number of `hyper-parameters', which are values which change the behaviour of the model (traditionally used to control over- and under-fitting). To approp

\subsubsection{Dataset Preparation}
\label{sub:data-prepare}

\begin{table}[!htbp]
    \caption{ Chosen UCI datasets }
    {\centering
    \footnotesize
    \begin{tabular}{m{.15\textwidth} | c  c  c | m{.5\textwidth}}

        Dataset & Shorthand & \#F\footnotemark[1] & \#E\footnotemark[2] & Description \\
        \hline

        Automobile & \emph{automobile} & 26 & 205 &
        Dataset consisting of details of various automobiles. The task is to predict the insurance category.
        Chosen because obvious grouping of attributes exists. \\
        %\url{https://archive.ics.uci.edu/ml/datasets/automobile} \\
        \hline
        Bank marketing \cite{moro2014data} & \emph{bank} & 17 & 45211 &
        Dataset consisting of details of customers of a banks. The task is to predict if the customer will respond to a marketing campaign.
        Chosen because obvious grouping of attributes exists.
        \\
        %\url{https://archive.ics.uci.edu/ml/datasets/Bank+Marketing} \\
        \hline
        Car evaluation & \emph{car} & 6 & 1728 &
        Synthetic dataset consisting of basic details of various automobiles (e.g. price, comfort). The task is to predict the overall `acceptability' of an automobile.
        Chosen because a specific hierarchy is given by the dataset constructors.
        \\
        %\url{https://archive.ics.uci.edu/ml/datasets/car+evaluation} \\
        \hline
        Contraceptive method choice & \emph{contraceptive} & 9 & 1473 &
        Dataset consisting of the details of various couples in Indonesia. The task is to predict the type of contraceptive method used by a couple.
        Chosen because obvious grouping of attributes exists.
        \\
        %\url{https://archive.ics.uci.edu/ml/datasets/Contraceptive+Method+Choice) \\
        \hline
        Mushroom & \emph{mushroom} & 22 & 8124 &
        Dataset consisting of details of various mushroom species. The task is to predict whether a given mushroom is poisonous or not.
        Chosen because obvious grouping of attributes exists.
        \\
        %\url{https://archive.ics.uci.edu/ml/datasets/mushroom} \\
        \hline
        Nursery evaluation & \emph{nursery} & 9 & 12960 &
        Dataset consisting of basic details of various nurseries. The task is to predict the overall `acceptability' of a nursery. Note: labels for the data are generated algorithmically.
        Chosen because a specific hierarchy is given by the dataset constructors.
        \\
        %\url{https://archive.ics.uci.edu/ml/datasets/nursery} \\
        \hline
        Seismic-bumps \cite{sikora2010application}& \emph{seismic} & 19 & 2584 &
        Dataset consisting of various readings from within a coal mine. The task is to predict if there will be a significant seismic event within the next 8 hours.
        Chosen because of the opportunity to use a JSON \emph{list} for certain data.
        \\
        %\url{https://archive.ics.uci.edu/ml/datasets/seismic-bumps} \\
        \hline
        Student performance \cite{cortez2008using} & \emph{student} & 33 & 649 &
        Dataset consisting of details of various students from two schools in Portugal, including prior academic results. The task is to predict the student's grade at the end of the year.
        Chosen because obvious grouping of attributes exists.
        \\
        %\url{https://archive.ics.uci.edu/ml/datasets/student\%2Bperformance} \\
        %\hline
    \end{tabular}}\\
    \footnotemark[1] Number of features/attributes \\
    \footnotemark[2] Number of examples

    \label{tab:uci_datasets}
\end{table}

For each of those benchmark datasets, we recover a hierarchical structure from its attributes, then represent its hierarchical structure in the JSON format. To do so, we apply a number of criteria to select different benchmark datasets for our experiments as follows:
1) Datasets where specific hierarchies are given explicitly (e.g. \emph{Car Evaluation});
2) Datasets where a hierarchy can be inferred either from variable names (e.g. \emph{Mushroom}) or domain knowledge (e.g. \emph{Automobile}); and
3) Datasets with a clear candidate for lists (e.g. \emph{Seismic}).
In summary, the information of all the chosen datasets are described in Table~\ref{tab:uci_datasets}.

There are two different representations for input attributes in each of the chosen datasets; feature vector and JSON format. For feature vector, the input of each example is converted into a vector of various features of the data. Categorical features are encoded as one-hot vectors, and numerical features were normalised by their mean and standard-deviation (across the whole dataset).
For JSON format, the input of each example is given as a semantic tree expressed in the JSON format. For each dataset, we manually construct a plausible JSON schema, and convert the input of each example automatically to a JSON entry by using this schema. Example entries for each dataset are shown in Appendix~\ref{ap:uci_examples}.

Those datasets with particularly unbalanced classes (i.e. where >80\% of examples were a single class) are re-balanced to a 2:1 split by randomly under-sampling on the dominant class, which enables use of classification accuracy across all datasets without involving the mix of classification accuracy issue and avoids using other evaluation metrics for unbalanced datasets.

\subsubsection{Main Experiments}
\label{sub:ex1}

In our comparative study, our JSON models, deep-set and LSTM based implementations described in Section~\ref{subsect:json-net}, are employed to work on the JSON format directly for supervised learning.
For the main neural network baseline, we used a Multi-Layer Perceptron (MLP) -- a fully-connected neural network architecture which accepts fixed-size feature vectors as inputs---applied to the standard categorical and numerical features for each dataset.
Furthermore, we compare to a uniquely existing JSON model developed in \cite{pevny2019universal}, which is referred to as the ``\emph{JSON-Grinder}" model, using the source code provided by the authors\footnote{This can be found at the following url: \url{https://github.com/pevnak/JsonGrinder.jl}}.
Note that this implementation only handles datasets with fixed JSON schemas, hence is only applicable to certain datasets.
In addition, we also employe a number of off-the-shelf machine learning models working on the feature vector representation, including logistic regression, support vector machine (SVM) and random forest. 

Both our JSON models, and the MLP were implemented using \emph{pytorch} and trained using the Adam optimiser \cite{kingma2014adam}, on the softmax cross-entropy loss of the raw output of each network.
Due to the difficulty of batching examples using the proposed framework, we used pseudo-minibatching, where gradients are accumulated over several examples, and then applied every few steps. This is mathematically equivalent, but generally less computationally efficient due to reduced parallelism.
For the JSON-Grinder model we used the original Julia code provided by the authors, while for the non-neural baselines we used the standard implementations from the \emph{sklearn} python library.

\begin{table}[!ht]
    \caption{ UCI datasets results }

    \centering
    \parbox{.8\textwidth}{\centering
    \begin{tabular}{c | r r r r  }
    Dataset & \multicolumn{1}{c}{MLP} & \multicolumn{1}{c}{JSON-Grinder $\dagger$} &
    \multicolumn{1}{c}{Set-based $\ddagger$} & \multicolumn{1}{c}{LSTM-based $\ddagger$} \\
    \hline
    automobile &  79.0\% (1.95) &\multicolumn{1}{c}{---}&  82.4\% (4.97) &  84.9\% (3.58)\\
    bank &  79.3\% (0.44) &  63.2\% (15.75)$^1$ &  78.9\% (0.60) &  79.3\% (0.42)\\
    car &  79.9\% (1.40) &\multicolumn{1}{c}{---}& 100.0\% (0.00) & 100.0\% (0.00)\\
    contraceptive &  55.3\% (2.78) &  55.5\% (5.17) &  54.0\% (5.03) &  53.4\% (2.31)\\
    mushroom & 100.0\% (0.00) & 100.0\% (0.00) &  99.7\% (0.37) & 100.0\% (0.00)\\
    nursery &  95.6\% (0.69) &\multicolumn{1}{c}{---}&  99.2\% (0.89) & 100.0\% (0.02)\\
    seismic &  73.3\% (5.28) &\multicolumn{1}{c}{---}&  72.9\% (3.37) &  71.6\% (3.86)\\
    student &  37.0\% (5.49) &  36.0\% (4.31) &  30.2\% (3.59) &  34.2\% (5.12)\\
    %student (reg) & 0.020 (0.003) &\multicolumn{1}{c}{---}& 0.020 (0.003) & 0.021 (0.004)\\
    \end{tabular}

    \begin{flushleft}
    $\dagger$ - Model of \cite{pevny2019universal} \\
    $\ddagger$ - Our proposed models \\
    $^1$ - Low average due to a low result on a single fold
    \end{flushleft}}
    \label{tab:json_uci_results}
\end{table}

Experiments are conducted by using 5-fold cross validation \cite{kohavi1995study} and taking the the mean and standard deviation of classification accuracy across the folds.
Hyper-parameter search is done independently for each fold using grid search on a held-out 10\% validation set, except for three small datasets (\emph{automobile}, \emph{seismic}, and \emph{student}) where $3$-fold cross-validation was applied on the whole train fold.
The chosen hyper-parameters for each fold are given in Tables \ref{tab:json_uci_parameters_mlp} to \ref{tab:json_uci_parameters_lstm}. An error in the hyper-parameter tuning code means that the learning rate is always set to the default setting, and not adjusted. However, tests with adjusting the learning rate on the \texttt{car} and \texttt{nursery} datasets did not lead to improved performance for the MLP model, hence, this does not appear to have adversely affected the baseline over the proposed models.
A subset of the results, including only the neural network models, is described in Table~\ref{tab:json_uci_results}. The full results is reported in Table~\ref{tab:full_json_uci_results}.

%% This paragraph needs re-writing after having all the new results.
It is evident from Tables \ref{tab:json_uci_results} and \ref{tab:full_json_uci_results} that our JSON models outperforms other models on several datasets where the semantic structure information plays an important role in classification, e.g, \emph{Automobile} and \emph{Car}. Overall, the performance yielded by our JSON models is comparable to other models on all the datasets used in our experiments.

\subsubsection{Ablation Study}

\begin{table}[]
    \caption{ Ablation study results on UCI datasets }
    \centering
    \begin{tabular}{c | r r r r  }
    Fraction & \multicolumn{1}{c}{LSTM-Based} & \multicolumn{1}{c}{Pathless} &
    \multicolumn{1}{c}{Homogenous} & \multicolumn{1}{c}{Both} \\
    \hline
    automobile &  84.9\% (3.58) &  70.7\% (7.07) &  86.3\% (7.33) &  77.6\% (3.90)\\
    bank &  79.3\% (0.42) &  76.9\% (1.00) &  79.7\% (0.73) &  79.2\% (0.69)\\
    car & 100.0\% (0.00) &  94.4\% (0.66) & 100.0\% (0.00) &  94.0\% (1.31)\\
    contraceptive &  53.4\% (2.31) &  45.6\% (3.36) &  52.0\% (4.18) &  49.4\% (2.05)\\
    mushroom & 100.0\% (0.00) & 100.0\% (0.00) & 100.0\% (0.00) & 100.0\% (0.03)\\
    nursery & 100.0\% (0.02) & 100.0\% (0.02) & 100.0\% (0.02) & 100.0\% (0.02)\\
    seismic &  71.6\% (3.86) &  67.4\% (3.35) &  68.9\% (4.82) &  66.9\% (2.89)\\
    student &  34.2\% (5.12) &  32.5\% (4.77) &  34.7\% (5.24) &  31.4\% (4.97)
    \end{tabular}
    \label{tab:json_ablation_results}
\end{table}

To investigate the importance of different components in our framework, we also carry out an ablation study where certain key components of the TreeLSTM-based model, corresponding to the various aspects of the framework, are removed, and the resultant simplified model is re-run on those UCI datasets with the same settings described in Section \ref{sub:ex1}. Thus, we have three simplified JSON models: 1) \emph{Homogenous} types, where all types within the same category were treated as a single type - objects for containers, and strings for primitives. In particular $r_{list}$ was replaced by $r_{object}$ and $r_{boolean}$ and $r_{number}$ were set to $r_{string} (\texttt{str}(x))$;
2) \emph{Pathless}, where the path information was removed from the network. This amounts to tying the weights for all values of path $p$; and 
3) \emph{Both} ablations, where both of the above were applied.

%% This paragraph needs re-writing after having all the new results.
Experimental results of this ablation study are listed in Table~\ref{tab:json_ablation_results}. The experimental results suggest that the main components in our framework play an irreplaceable role apart from the embedding function for data of different primitive types.

\subsubsection{Data Efficiency}
\label{subsubsec:uci_experiments_data_efficiency}

\begin{table}[!htbp]
    \caption{ Data efficiency results on fractions of Poker Hands dataset }
    \centering
    \parbox{.85\textwidth}{\centering
    \begin{tabular}{c | r r r r r  }
    Fraction & \multicolumn{1}{c}{MLP} & \multicolumn{1}{c}{JSON-Grinder $\dagger$} &
    \multicolumn{1}{c}{Set-based $\ddagger$} & \multicolumn{1}{c}{LSTM-based $\ddagger$} &
    \multicolumn{1}{c}{Tailored $\ddagger$*}\\
    \hline
    5\% &  47.7\% (0.68) &  80.6\% (5.05) &  65.0\% (4.70) &  53.6\% (1.72) &  96.1\% (1.4) \\
    20\% &  82.7\% (9.47) &  97.9\% (0.46) &  94.4\% (1.97) &  79.8\% (6.57) &  96.9\% (0.4) \\
    50\% &  86.3\% (6.93) &  99.1\% (0.19) &  95.6\% (3.43) &  98.4\% (0.23) &  97.4\% (0.1) \\
    100\% &  97.6\% (0.67) &  86.4\% (7.72)$^1$ &  97.5\% (0.25) &  98.4\% (0.21) &  97.6\% (0.1) \\
    \end{tabular}

    \begin{flushleft}
    $\dagger$ - Model of \cite{pevny2019universal} \\
    $\ddagger$ - Our proposed models \\
    * - Model of Section~\ref{subsubsec:tailored-experiments} \\
    $^1$ - Low average due to a low result on a single run
    \end{flushleft}}
    \label{tab:json_poker_results}
\end{table}

In order to test the effect of using different amounts of training data in learning models, we conduct a data-efficiency study where the training set size was reduced to various fractions.
For this study the UCI \emph{poker hands} dataset is used due to its large number of training examples. This dataset is preprocessed with the input of feature-vector and the JSON format in the same fashion as those UCI datasets described in Section \ref{sub:data-prepare}.
Each learning model is run with training sets of 5\%, 20\%, 50\%, and 100\% of the original number of training examples, where fractional training datasets were constructed by randomly under-sampling from the original training set.
Hyper-parameter tuning was done for each model/fraction using random-search on random splits of the train data, using the remaining train examples as the validation set.
Testing was done on the provided test set, and each model was trained 5 times using different splits of the full training set. We report the mean, and standard deviation, of the classification accuracy on the test set over the 5 runs.
Results are given in Table~\ref{tab:json_poker_results}.
Chosen hyper-parameters are given in Table~\ref{tab:json_uci_parameters_poker}.

\subsubsection{Path-specific functions}\label{subsubsec:tailored-experiments}

To demonstrate how the path specific functions of Section~\ref{subsubsect:path-specific-funcs} can be used to tailor an implementation of the framework to a particular dataset, we repeated the experiments on the \emph{poker hands} dataset with a JSON implementation of the STRLA framework employing path-specific functions.
We use the mapping dictionary of Figure~\ref{fig:poker_path_overrides}, where \texttt{sumTreeLSTM}, \texttt{LSTM}, and \texttt{deepSets} correspond to the same container embedding functions described in Section~\ref{subsect:json-container}, and \texttt{catEmbedding} corresponds to a categorical embedding function, where each distinct value is mapped to a separate latent vector.
The resulting model is relatively simpler than the Set-based and TreeLSTM-based exemplar models, while the path-specific functions allow the permutation-invariance of the cards in the hand to be targeted via a permutation-invariant function.
The results for this ``Tailored'' model are shown in the final column of Table~\ref{tab:json_poker_results}.

\begin{figure}[h]
    \centering
\begin{lstlisting}
{
    'object~..': 'sumTreeLSTM',  'array~..': 'LSTM',
    'string~..': 'catEmbedding', 'number~..': 'catEmbedding',

    '.*': 'deepSets'
}
\end{lstlisting}

    \caption{Path mapping dictionary used for the \emph{poker-hands} dataset}
    \label{fig:poker_path_overrides}
\end{figure}

\subsection{Reinforcement learning task}
\label{sub:rl-task}

We also demonstrate our approach on a toy reinforcement learning task consisting of navigating an avatar on a grid-world to collect various boxes. Each time the environment is reset, between 1 to 4 boxes are randomly placed on a 9$\times$9 grid, and the avatar is reset to the centre square. One point is obtained for each box the agent collects by navigating to, and the environment is reset when there are no boxes remaining, or after 100 steps.
%Generate JSON descriptions by applying JSON parser to underlying python object hierarchy
We implemented this task in py-vgdl \cite{schaul2013video}, which was adapted to use a modified OpenAI gym interface \cite{brockman2016openai}.
To obtain JSON state-descriptions, we encode the underlying python objects of each game object into corresponding JSON objects using the in-built python JSON parser, where we encode objects by using the \texttt{\_\_dict\_\_} method to get a dict of attributes of each object. These object-level JSON descriptions are compiled into a JSON list to get a single JSON description of the game state.

\begin{figure}[h]
    \centering
\begin{lstlisting}
{ 'array~..': 'LSTM',
'object~..': 'sum',
'string~..': 'categoricalEmbedding',
'number~..': 'categoricalEmbedding',

'.*': 'deepSetsEquivariant',
'number~..rect..': 'embedNumber',

'..lastrect': 'ignore',
'..physicstype': 'ignore',
'..physics': 'ignore',
'..img': 'ignore',
'..color': 'ignore',
'..image': 'ignore',
'..scale_image': 'ignore',
'..stypes': 'ignore' }
\end{lstlisting}
    \caption{Path mapping dictionary used for py-VGDL}
    \label{fig:vgdl_path_overrides}
\end{figure}

%use PPO algorithm
%hyper-parameters
% use network of woof & chen

We use the fully-observed (i.e. without history based policy) Proximal Policy Optimisation algorithm \cite{schulman2017proximal}, using our JSON network model to encode the environment state into a single latent vector, which we branch off into separate policy and value heads using a single linear transformation.
Since the game-state consists of multiple object-level descriptions, for encoding this final list we use the architecture used in \cite{woof2018learning} (which is a variant of the architecture of \cite{zaheer2017setnet}), which is applied via path-specific function mapping. Additionally, since various object attributes are cosmetic, or otherwise irrelevant, we ignore these via path-specific function mappings. The full list of function mappings is given in Figure~\ref{fig:vgdl_path_overrides}.

For our baseline we use the object-based method of \cite{woof2018learning} using the same object-level features, but adapted for the PPO algorithm. This uses the same root level architecture, but with hand-defined object-level feature vectors. We also compare to an image-based agent using a Convolutional Neural Network architecture, where input is given as a render of the game-screen. The same hyper-parameters settings were used for all approaches.

\begin{figure}[!htbp]
    \centering %TODO: Update figure
    \includegraphics[width=.5\textwidth]{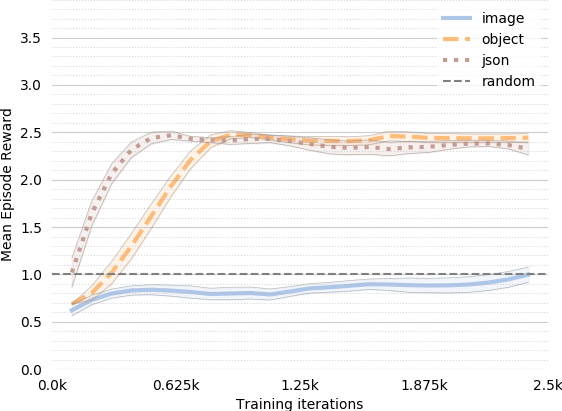}
    \caption{Mean average episode reward over 4 runs for the JSON-based agent and the object-based agent. Shaded region denotes the region within 1 standard error of the average over 4 runs (approx 70\% confidence interval). Results are smoothed for readability using a Savitzky-Golay filter with window size 11, polynomial degree 3.}
    \label{fig:json_agents_results_procgen}
\end{figure}

Results are given in Figure~\ref{fig:json_agents_results_procgen}. It should be noted that the training speed of the JSON-based agent was substantially slower (by about 10 times) than the baseline agent.

\section{Discussion}
\label{sect:discuss}

%This section needs re-writing; most of the analysis should be moved to the experiment section but implication and issues arising from this study need to be enhanced significantly. 

The experimental results suggest that the proposed approach is competitive with other approaches, and even exceeds the performance of other models on certain datasets.
In particular on the \emph{car} and \emph{nursery} datasets, both our exemplar JSON models attain near perfect accuracy, while the other models fall short.
Both of these datasets are synthetic hierarchical datasets, originally intended to test the hierarchical reasoning powers of expert systems. Hence, this suggests that the proposed framework is able to take advantage of hierarchy in data where it exists.
The two proposed models also performed comparably to the JSON-Grinder model, with the exception of the \emph{poker-hands} dataset, where the relative simplicity of the underlying JSON-Grinder architecture likely helped it avoid overfitting on the smaller dataset fractions.
However, the JSON-Grinder model was only applicable to those datasets where the dataset schema was fixed, and couldn't handle examples with missing attributes or where certain attributes changed types across different data examples, highlighting the advantage of the dynamic architecture construction of the proposed STRLA framework.
In addition to the \emph{car} and \emph{nursery} datasets, all the JSON models (including the JSON-Grinder model of \cite{pevny2019universal}) proved to be much more data-efficient on the \emph{poker-hands} dataset. This was likely down to the use of permutation invariant functions for representing the cards in the hand in the Set-based and JSON-Grinder models. Meanwhile the MLP baseline uses a fully connected layer with inputs for 5 cards, meaning each permutation of cards is treated separately, increasing the number of different cases the model was required to learn.
This was further demonstrated by the tailored model, which used path specific embedding functions to specifically apply a permutation-invariant function to the constituent cards.%, resulting in a model which was able to achieve a high classification accuracy on only

However, a significant downside of the proposed approach was its slow iteration-speed compared to other methods. While some of this was down to the greater complexity of the underlying constructed neural network architectures, low CPU/GPU utilisation during training suggests that slow training speeds were largely down to inefficiencies caused by bottlenecks in the framework code.
Unfortunately, the use of dynamic construction of the network architecture means that the framework difficult to optimise. However, more careful integration of the framework code with the underlying deep learning framework could help alleviate these bottlenecks.

% What about ablation, why homogenous not needed
% - Very few 'lists' in data.
The importance of differentiating elements by path information was clearly demonstrated in the ablation study, where the pathless model attained lower accuracy across all datasets.
However, the same was not true for the use of separate embedding functions for each type, as the model with homogenous embedding functions performed very similarly to the original model.
On the other hand, the difference in performance between the Set-based and LSTM-based models on the \emph{poker hands} data efficiency study suggests that the choice of components matters, at least in certain cases.
This suggests that the appropriate choice of function may be less dependent on the specific element type, and more on the actual data of the element itself.
While theoretically JSON may only have three primitive types, in practice, JSON elements can take many different forms (for example \emph{string}s can be used contain plain-text, dates, encoded binary data, etc.),
hence using the same function $f_t$ for all instances of a particular type may not be appropriate.
%This is also true for other data formats
%Additionally, the hierarchicality of the data will also vary.
%One way this could be addressed is via path-specific function mappings, where instead of a single embedding function $f_t$ for all elements of the same type, instead multiple functions $\{f_{t,i}\}_{i=1...k}$ could be used, where the choice of function $f_{t,i}$ is dependent on the element path.
This can be addressed using the path-specific functions of Section~\ref{subsubsect:path-specific-weights};
by using path-specific functions, the most appropriate choice of embedding function can be used for each element of the data hierarchy rather than relying on the same embedding function for all elements of the same type.
However, the appropriate function for each path has to be determined by hand. This reintroduces an unwanted reliance on human input into application of the framework.
For future work, one approach would be apply a form of neural architecture search method \cite{elsken2018neural}, where various choices of function mapping for each path are tested, keeping the best-performing functions.

%Many of these formats can be detected by inspecting the data (for example, "12:43 08/10/2018" is clearly a time and a date), hence a database could be (manually) constructed based on common data formats, along with override functions for the given format.

In Section~\ref{subsect:e-path} we describe two different simple approaches to incorporating path information into the embedding functions \{$f_p$\} and \{$f_c$\}, which we use in our exemplar implementations.
While easy to implement, these approaches treat paths as discrete identifiers, which ignores a large amount of nuances about paths that could otherwise be exploited.
In particular, names in JSON and other semantic tree-structures often comprise of natural language descriptions of the data within; if the names of two elements are similar, then the elements are usually semantically similar.
If the semantic related between paths could be better exploited then potentially some learning could be shared between the functions of elements of different, but similar, paths. That is, an update to $f_t(\cdot, p_{e_1})$ could also update $f_t(\cdot, p_{e_2})$, where $p_{e_1}$ and $p_{e_2}$ are semantically related.
This could be used in transfer learning, where functions $f_t(\cdot, p_e)$ trained on a source dataset could be applied directly to a target dataset, even when the paths of the elements in the different datasets are not exactly identical.
%There are many possible ways this could be implemented...

%As highlighted in previous sections, the element path provides the important contextual information, hence the proper use of element path in our framework leads to the considerable performance gain. Likewise,  we can also define the description of a given element in an STS. Given an element $e$ within a parent element $e_0$, if $e$ is directly wrapped, then the description $d_e$ of $e$ is the description of this wrapper (which may be $\emptyset$), otherwise $d_e := \emptyset$. An element is directly wrapped only if its most immediate wrapper wraps only itself; an element whose most immediate wrapper wraps a container of that element is not directly wrapped. We believe that introducing this concept would be useful in capturing the useful information conveyed in semantic tree-structured data. The usefulness of element descriptions can be explored in an alternative implementation.

%There are numerous ways that path information could be incorporated into functions $f_*$. However, the framework above still provides plenty of scope to explore alternative ways of incorporating path information.

\section{Conclusion}
\label{sect:conclude}

In this paper we have presented a novel framework for applying end-to-end learning on semantic tree-structured data format such as JSON and XML. We have presented exemplar instantiations of this framework for the JSON format, and evaluated them on a variety of different datasets. Our results demonstrate that the proposed models are comparable with existing approaches based on hand-engineered features, even exceeding them on datasets with innate hierarchicality, while being directly applicable to arbitrary JSON data without requiring additional human intervention.
While there are still a number of outstanding issues with our implementations of this framework, we believe that these initial results demonstrate the feasibility of our approach, and the end-to-end learning on semantic tree-structured data is promising for AutoML and worth exploring and developing further. % with implications for auto-ml

In future work, we would apply our framework to diversified end-to-end learning tasks on semantic tree-structured data by exploring those proposed improvements to our implementation described in Section~\ref{sect:discuss}. Moveover, we are going to explore implementations of our framework for further formats such as XML, HTML, YAML, etc.



%% References


\bibliographystyle{elsarticle-num}
\bibliography{bibliography}

\begin{thebibliography}{10}
\expandafter\ifx\csname url\endcsname\relax
  \def\url#1{\texttt{#1}}\fi
\expandafter\ifx\csname urlprefix\endcsname\relax\def\urlprefix{URL }\fi
\expandafter\ifx\csname href\endcsname\relax
  \def\href#1#2{#2} \def\path#1{#1}\fi

\bibitem{zaheer2017setnet}
M.~Zaheer, S.~Kottur, S.~Ravanbakhsh, B.~Poczos, R.~Salakhutdinov, A.~Smola,
  Deep sets, in: Advances in Neural Information Processing Systems, 2017, pp.
  3391--3401.

\bibitem{zhou2018graphnet}
J.~Zhou, C.~Cui, Z.~Zhang, C.~Yang, Z.~Liu, L.~Wang, C.~Li, M.~Sun, Graph
  neural networks: A review of methods and applications, arXiv preprint
  arXiv:1812.08434 (2018).

\bibitem{HAMMER20041061}
B.~Hammer, A.~Micheli, A.~Sperduti, M.~Strickert, Recursive self-organizing
  network models, Neural Networks 17~(8) (2004) 1061 -- 1085.

\bibitem{zhang2018t2v}
H.~Zhang, S.~Wang, X.~Xu, T.~Chow, Q.~Wu, {Tree2Vector}: Learning a vectorial
  representation for tree-structured data, IEEE Transactions on Neural Networks
  and Learning Systems 29~(11) (2018) 5304--5318.

\bibitem{socher2013treebank}
R.~Socher, A.~Perelygin, J.~Wu, J.~Chuang, C.~Manning, A.~Ng, C.~Potts,
  Recursive deep models for semantic compositionality over a sentiment
  treebank, in: Proceedings of the Conference on Empirical Methods in Natural
  Language Processing, 2013, pp. 1631--1642.

\bibitem{irsoy2013bidirectional}
O.~Irsoy, C.~Cardie, Bidirectional recursive neural networks for token-level
  labeling with structure, arXiv preprint arXiv:1312.0493 (2013).

\bibitem{paulus2014belief}
R.~Paulus, R.~Socher, C.~Manning, Global belief recursive neural networks, in:
  Advances in Neural Information Processing Systems, 2014, pp. 2888--2896.

\bibitem{irsoy2014deeprecursive}
O.~Irsoy, C.~Cardie, Deep recursive neural networks for compositionality in
  language, in: Advances in Neural Information Processing Systems, 2014, pp.
  2096--2104.

\bibitem{tai2015improved}
K.~Tai, R.~Socher, C.~Manning, Improved semantic representations from
  tree-structured long short-term memory networks, in: Proceedings of ACL,
  Vol.~1, 2015, pp. 1556--1566.

\bibitem{kokkinos2017tree}
F.~Kokkinos, A.~Potamianos, Structural attention neural networks for improved
  sentiment analysis, in: Proceedings of the Conference of the ACL European
  Chapter, 2017, pp. 586--591.

\bibitem{pevny2019universal}
T.~Pevny, V.~Kovarik, Approximation capability of neural networks on spaces of
  probability measures and tree-structured domains, arXiv preprint
  arXiv:1906.00764 (2019).

\bibitem{hochreiter1997long}
S.~Hochreiter, J.~Schmidhuber, Long short-term memory, Neural computation 9~(8)
  (1997) 1735--1780.

\bibitem{Dua:2019}
D.~Dua, C.~Graff, \href{http://archive.ics.uci.edu/ml}{{UCI} machine learning
  repository} (2017).
\newline\urlprefix\url{http://archive.ics.uci.edu/ml}

\bibitem{AutoML}
{AutoML Freiburg}, \url{https://www.automl.org/}, accessed: 2019-07-20.

\bibitem{coller1996task-dependent}
C.~Coller, A.~Kuchler, Task-dependent distributed structure-representations by
  backpropagation through structure, in: International Conference on Neural
  Networks, IEEE, 1996, pp. 347--352.

\bibitem{frasconi1998general}
P.~Frasconi, M.~Gori, A.~Sperduti, A general framework for adaptive processing
  of data structures, IEEE Transactions on Neural Networks 9~(5) (1998)
  768--786.

\bibitem{socher2011parsing}
R.~Socher, C.~Lin, A.~Ng, C.~Manning, Parsing natural scenes and natural
  language with recursive neural networks, in: International Conference on
  Machine Learning, 2011, pp. 129--136.

\bibitem{rabinovich2017abstract}
M.~Rabinovich, M.~Stern, D.~Klein, Abstract syntax networks for code generation
  and semantic parsing, in: Proceedings of ACL, 2017, pp. 1139--1149.

\bibitem{chen2018t2t}
X.~Chen, C.~Liu, D.~Song, Tree-to-tree neural networks for program translation,
  in: Advances in Neural Information Processing Systems, 2018, pp. 2547--2557.

\bibitem{li2015tree}
J.~Li, T.~Luong, D.~Jurafsky, E.~Hovy, When are tree structures necessary for
  deep learning of representations?, in: Proceedings of the Conference on
  Empirical Methods in Natural Language Processing, 2015, pp. 2304--2314.

\bibitem{Knuth:1964:BNF}
D.~Knuth, Backus normal form vs. {Backus Naur} form, Commun. ACM 7~(12) (1964)
  735--736.

\bibitem{clark1999xml}
J.~Clark, S.~DeRose, et~al., Xml path language (xpath) version 1.0 (1999).

\bibitem{goessner2007jsonpath}
S.~Goessner, Jsonpath-xpath for json, URL http://goessner.
  net/articles/JsonPath (2007).

\bibitem{dua2019uci}
D.~Dua, C.~Graff, \href{http://archive.ics.uci.edu/ml}{{UCI} machine learning
  repository} (2017).
\newline\urlprefix\url{http://archive.ics.uci.edu/ml}

\bibitem{moro2014data}
S.~Moro, P.~Cortez, P.~Rita, A data-driven approach to predict the success of
  bank telemarketing, Decision Support Systems 62 (2014) 22--31.

\bibitem{sikora2010application}
M.~Sikora, et~al., Application of rule induction algorithms for analysis of
  data collected by seismic hazard monitoring systems in coal mines, Archives
  of Mining Sciences 55~(1) (2010) 91--114.

\bibitem{cortez2008using}
P.~Cortez, A.~M.~G. Silva, Using data mining to predict secondary school
  student performance (2008).

\bibitem{kingma2014adam}
D.~P. Kingma, J.~Ba, Adam: A method for stochastic optimization, arXiv preprint
  arXiv:1412.6980 (2014).

\bibitem{kohavi1995study}
R.~Kohavi, et~al., A study of cross-validation and bootstrap for accuracy
  estimation and model selection, in: Ijcai, Vol.~14, Montreal, Canada, 1995,
  pp. 1137--1145.

\bibitem{schaul2013video}
T.~Schaul, A video game description language for model-based or interactive
  learning, in: Computational Intelligence in Games (CIG), 2013 IEEE Conference
  on, IEEE, 2013, pp. 1--8.

\bibitem{brockman2016openai}
G.~Brockman, V.~Cheung, L.~Pettersson, J.~Schneider, J.~Schulman, J.~Tang,
  W.~Zaremba, Openai gym, arXiv preprint arXiv:1606.01540 (2016).

\bibitem{schulman2017proximal}
J.~Schulman, F.~Wolski, P.~Dhariwal, A.~Radford, O.~Klimov, Proximal policy
  optimization algorithms, arXiv preprint arXiv:1707.06347 (2017).

\bibitem{woof2018learning}
W.~Woof, K.~Chen, Learning to play general video-games via an object embedding
  network, in: 2018 IEEE Conference on Computational Intelligence and Games
  (CIG), IEEE, 2018, pp. 1--8.

\bibitem{elsken2018neural}
T.~Elsken, J.~H. Metzen, F.~Hutter, Neural architecture search: A survey, arXiv
  preprint arXiv:1808.05377 (2018).

\end{thebibliography}


\newpage
\begin{appendices}
\renewcommand{\theequation}{A.\arabic{equation}}
\renewcommand{\thetable}{A.\arabic{table}}
\renewcommand{\thefigure}{A.\arabic{figure}}
\setcounter{equation}{0}
\setcounter{table}{0}
\setcounter{figure}{0}

%\renewcommand{\thealgorithm}{A.\arabic{algorithm}}
%\setcounter{algorithm}{0}

%%%%%%%%%%%%%%%%%% Appendix A %%%%%%%%%%%%%%%%%%%%%%%%%%%%%%%%%%%%%%%%%%%%%%%%%%%%%%%%%

\section{JSON and XML Specification}
\label{appA}

In this appendix, we describe the specification of JSON and XML formats based on the meta-notational scheme described in Section~\ref{subsect:notation}.

Figure~\ref{fig:json} illustrates the specification of the JSON format with the BNF syntax, which covers all the permissible elements in JSON. and~\ref{fig:xml}, respectively.

\begin{figure*}[h]
    \centering
\begin{lstlisting}
<json> ::= <primitive> | <container>

<primitive> ::= <number> | <string> | <boolean>
; Where:
; <number> is a valid real number expressed in one of a number of given formats
; <string> is a string of valid characters enclosed in quotes
; <boolean> is one of the literal strings 'true', 'false', or 'null' (unquoted)

<container> ::= <object> | <array>
<array> ::= '[' [ <json> *(', ' <json>) ] ']' ; A sequence of JSON values
<value> ::= <json> ; A transparent wrapper
<object> ::= '{' [ <member> *(', ' <member>) ] '}' ; A sequence of 'members'
<member> ::= <name> ': ' <json> ; A pair consisting of a name, and a JSON value
\end{lstlisting}
    \caption{Specification of JSON format with the BNF syntax. }
    \label{fig:json}
\end{figure*}

In general, XML consists of many elements. Some elements such as CD data, processing instructions, header and comments are less semantically informative from a supervised learning perspective. Therefore, we ignore such elements and our description focuses on only the main elements in XML. As a result, Figure~\ref{fig:xml} illustrates the specification of the core XML format with the BNF syntax.

\begin{figure*}[h]
    \centering
\begin{lstlisting}
<xml-document> ::= <header> <xml>

<xml> ::= <primitive> | <container> | <entity-reference>

<primitive> ::= <string> ; Only a single primitive type
<container> ::= *(<tagged> | <empty-tag>) ; Only a single container type

<tagged> ::= '<' <name>  <attributes> '>' <xml> '</' <name> '>'
<empty-tag> ::= '<' <name>  <attributes> '/>' ; We treat this as <tag></tag>
; XML tag is a wrapper name and additional description data in the form of attributes
<attributes> ::= *(<string> '="' <string> '"')
\end{lstlisting}
    \caption{Specification of reduced XML format with the BNF syntax. }
    \label{fig:xml}
\end{figure*}

\section{LSTM and Tree-LSTM Networks}
\label{appB}

In this appendix, we describe two building blocks, LSTM \cite{hochreiter1997long} and Tree-LSTM \cite{tai2015improved}, used in the implementations of our framework.

\subsection{LSTM}
\label{sec:LSTM_eqns}

In the LSTM architecture \cite{hochreiter1997long}, an LSTM unit at each time step $t$ consists of several different mechanisms in the vector form including an input gate, ${\pmb i}_{t}$, a forget gate, ${\pmb f}_{t}$, an output gate, ${\pmb o}_{t}$, a memory cell, ${\pmb c}^{\rm LSTM}_{t}$ and a hidden state, ${\pmb h}^{\rm LSTM}_{t}$. The LSTM unit works with the transition equations as follows:
\begin{align*}
  & {\pmb i}_{t} = \sigma \bigg ( W^{(i)}{\pmb x}_t + V^{(i)}{\pmb h}_{t-1}
  + {\pmb b}^{(i)} \bigg ), &
  & {\pmb f}_{t} = \sigma \bigg ( W^{(f)}{\pmb x}_t + V^{(f)}{\pmb h}_{t-1}
  + {\pmb b}^{(f)} \bigg ),    \\
  & {\pmb o}_{t} = \sigma \bigg ( W^{(o)}{\pmb x}_t + V^{(o)}{\pmb h}_{t-1}
  + {\pmb b}^{(o)} \bigg ), &
  & {\pmb u}_{t} = \tanh \bigg ( W^{(u)}{\pmb x}_t + V^{(u)}{\pmb h}_{t-1}
  + {\pmb b}^{(u)} \bigg ),\\
   % & {\pmb c}^{\rm LSTM}_{t} = {\pmb i}_{t} \odot \mathbf u_{t} + {\pmb f}_{t} \odot {\pmb c}_{t-1}, &
  & {\pmb c}_{t} = {\pmb i}_{t} \odot \mathbf u_{t} + {\pmb f}_{t} \odot {\pmb c}_{t-1}, &
  & {\pmb h}^{\rm LSTM}_{t} = {\pmb o}_{t} \odot \tanh \bigg ( {\pmb c}^{\rm LSTM}_{t} \bigg ),
\end{align*}
where ${\pmb x}_t$ is the current input at time step $t$, $\odot$ denotes the elementwise multiplication operator, $\sigma(\circ)$ and $\tanh(\circ)$ are the standard sigmoid and the hyperbolic tangent functions, respectively.

Based on those transition equations described above, we can obtain the memory cell and hidden state vectors of the LSTM unit after a sequence,  $\{{\pmb x}_1, \cdots , {\pmb x}_k \bigg \}$, is processed:
\begin{equation*}
 {\pmb h}^{\rm LSTM}_k
 \Leftarrow {\rm LSTM} \bigg (\big \{{\pmb x}_1, \cdots , {\pmb x}_k \big \}, \Theta_{LSTM} \bigg ),
\end{equation*}
%\begin{equation*}
%  \bigg ({\pmb c}^{\rm LSTM}_k, {\pmb h}^{\rm LSTM}_k \bigg )
% \Leftarrow {\rm LSTM} \bigg (\big \{{\pmb x}_1, \cdots , {\pmb x}_k \big \}, \Theta_{LSTM} \bigg ),
%\end{equation*}
where $\Theta_{LSTM}$ is a collective notation of all the parameters, $W$s, $V$s and ${\pmb b}$s, in the LSTM unit.

\subsection{Tree-LSTM}
\label{sec:SumLSTM_eqns}

In general, the LSTM architecture \cite{hochreiter1997long} is limited to strictly sequential information propagation. To overcome this limitation, two extensions of the original LSTM architecture, \emph{Child-Sum} and \emph{N-ary} Tree-LSTMs have been made in \cite{tai2015improved} for exploiting the structural and the semantic information underlying syntactic parse trees. Here, we adapt one of two extensions, Child-Sum Tree-LSTM, for a build block used in our implementation of the container embedding functions. Unlike a syntactic parse tree, there is no external input to branch nodes in any semantic tree defined in Section \ref{sect:stree}. Therefore, the external input in the Child-Sum Tree-LSTM architecture is omitted in our description below.

Likewise, a \emph{Child-Sum Tree-LSTM} (SumLSTM) unit consists of the same mechanisms in the LSTM unit as described in Section \ref{sec:LSTM_eqns}. A main difference is that all the gates or cell receive the a sum of children's hidden states in the SumLSTM instead of a single one in the LSTM. Assume that there are $n_e$ children linked to SumLSTM unit $j$, and let $\tilde{\pmb h}_j$ denote the sum of its children's hidden states where $\tilde{\pmb h}_j = \sum_{n=1}^{n_e}{\pmb h}_n$.
With the same notation used in Section \ref{sec:LSTM_eqns}, the SumLSTM unit works with the transition equations  as follows:
\begin{align*}
 & {\pmb i}_{j} = \sigma \bigg ( V^{(i)}\tilde{\pmb h}_j
  + {\pmb b}^{(i)} \bigg ), &
  & {\pmb f}_{\!jn} = \sigma \bigg ( V^{(f)}{\pmb h}_{n}   + {\pmb b}^{(f)} \bigg ),    \\
  & {\pmb o}_{j} = \sigma \bigg ( V^{(o)}\tilde{\pmb h}_j
  + {\pmb b}^{(o)} \bigg ), &
  & {\pmb u}_{j} = \tanh \bigg ( V^{(u)}\tilde{\pmb h}_j
  + {\pmb b}^{(u)} \bigg ),\\
  & {\pmb c}^{\rm SumLSTM}_{j} = {\pmb i}_{j} \odot \mathbf u_{j} + \sum_{n=1}^{n_e}{\pmb f}_{\!jn} \odot {\pmb c}_{n}, &
  & {\pmb h}^{\rm SumLSTM}_{j} = {\pmb o}_{j} \odot \tanh \bigg ( {\pmb c}^{\rm SumLSTM}_{j} \bigg ),
\end{align*}

Based on those transition equations described above, we can obtain the memory cell and hidden state vectors of SumLSTM unit $j$ after all the children's memory cell and hidden state vectors, $\big \{({\pmb c}_1, {\pmb h}_1), \cdots , ({\pmb c}_{n_e}, {\pmb h}_{n_e} ) \big \}$, are processed:
\begin{equation*}
  \bigg ({\pmb c}^{\rm SumLSTM}_j, {\pmb h}^{\rm SumLSTM}_j \bigg )
 \Leftarrow {\rm LSTM} \bigg (\big \{({\pmb c}_1, {\pmb h}_1), \cdots , ({\pmb c}_{n_e}, {\pmb h}_{n_e} ) \big \}, \Theta_{SumLSTM} \bigg ),
\end{equation*}
where $\Theta_{SumLSTM}$ is a collective notation of all the parameters, $V$s and ${\pmb b}$s, in the SumLSTM unit. It is worth clarifying that unlike the LSTM, a Tree-LSTM unit requires both memory cell and hidden state vectors of its children's, $\big \{({\pmb c}_1, {\pmb h}_1), \cdots , ({\pmb c}_{n_e}, {\pmb h}_{n_e} ) \big \}$.

\section{XML Recursive Networks}
\label{ap:xml-net}

In this appendix we describe an exmplar implementation for the XML format as described in Figure~\ref{fig:xml}.

As illustrated in Figure \ref{fig:xml}, XML consists of three components: \emph{primitive}, \emph{container} and \emph{entity-reference}. As the entity-reference type may are semantically irrelevant to a supervised learning task, we consider only the primitive and the container components for the XML format.

In the XML, the primitive component has only one element: \emph{string}. Therefore, we can use the exactly same method presented in Section \ref{subsect:json-primitive} to carry out the string embedding function for the XML format (see Eq.(\ref{eq:json-str}) and the relevant description for details).

In the XML, there is only one container element: the \emph{tagged} element. As illustrated in Figures \ref{fig:json} and \ref{fig:xml}, however, the container elements in the JSON and XML differ as the XML wrapper provides an option for describing a branch node with a list of attributes. %Also, the tagged elements may be arranged in a specific order.
Thus, we can adapt our implementation of the JSON's array embedding function for this requirement. Here, we present an implementation of this container embedding function for the XML format by making an extension of the LSTM-based method presented in Section \ref{subsect:json-container}. For a branch node, $e$, it has element path, $p_e \in \mathcal P$, $n_e$ children and a list of $m$ attributes, $\big \{ {\rm id}_1\!=\! v_1, \cdots, {\rm id}_{m}\!=\! v_{m} \big \}$. As each attribute value, $v_j$, is a string for $j=1,\cdots,m$, we can use the same method described Section \ref{subsect:json-primitive} to encode such information as follows:
\begin{equation}
\label{eq:xml-atr}
  f_{atr}(v_j|\Theta_{atr,id_j}) =
    \frac{1}{|v_j|}\sum_{i=1}^{|v_j|} {\pmb h}^{\rm LSTM}_i,
\end{equation}
where $\Theta_{atr,id_j}$ is the collective notation of all the learnable parameters in the LSTM network (see Appendix \ref{sec:LSTM_eqns} for details) for attribute $j$  and ${\pmb h}^{\rm LSTM}_i$ is the hidden state vector when character $i$ is sequentially input to the LSTM network, which is obtained by
$$
%\bigg ({\pmb c}^{\rm LSTM}_i, {\pmb h}^{\rm LSTM}_i \bigg )
 {\pmb h}^{\rm LSTM}_i
 \Leftarrow {\rm LSTM} \bigg (\bigg \{{\rm lin}(v_{j1},\theta_{id_j}),\cdots,{\rm lin}(v_{ji},\theta_{id_j}) \bigg \}, \Theta_{atr,id_j} \bigg ),
$$
where $\theta_{id_j}$ is the collective notation of the parameters in the linear transformation.
Thus, the \emph{tag} embedding function is carried out with another LSTM as follows:
\begin{equation}
  f_{tag} \bigg ( \big \{{\pmb h}_1, \cdots,  {\pmb h}_{n_e} \big \}, \big \{ {\rm id}_1\!=\! v_1, \cdots, {\rm id}_{m}\!=\! v_{m} \big \}| \Theta_{tag,p_e} \bigg ) =
     {\pmb h}^{\rm LSTM}_{n_e} +
    \sum_{j=1}^m f_{atr}(v_j|\Theta_{atr,id_j}),
\end{equation}
%\begin{equation}
%  f_{tag} \bigg ( \big \{({\pmb c}_1, {\pmb h}_1), \cdots, ({\pmb c}_{n_e}, {\pmb h}_{n_e}) \big \}, A| \Theta_{tag,p_e} \bigg ) =
%    \bigg ( {\pmb c}^{\rm LSTM}_{n_e}, {\pmb h}^{\rm LSTM}_{n_e} +
%    \sum_{j=1}^m f_{atr}(v_j|\Theta_{atr,id_j}) \bigg ),
%\end{equation}
where $\Theta_{tag,p_e}$ is the collective notation of all the parameters in this LSTM for the element path, $p_e \in \mathcal P$.
%${\pmb c}^{\rm LSTM}_{n_e}$ and
${\pmb h}^{\rm LSTM}_{n_e}$ is achieved recurrently via
\begin{equation*}
  %\bigg ({\pmb c}^{\rm LSTM}_k, {\pmb h}^{\rm LSTM}_k \bigg )
  {\pmb h}^{\rm LSTM}_k
 \Leftarrow {\rm LSTM} \bigg (\big \{{\pmb h}_1, \cdots , {\pmb h}_k \big \}, \Theta_{LSTM} \bigg ),
\end{equation*}
where ${\pmb h}_k$ is the latent representation of child $k$.

\newpage
\section{Full results for main UCI experiments}

\begin{table}[!htbp]
    \centering

    \begin{tabular}{l | r r r r }

& \multicolumn{1}{c}{MLP} & \multicolumn{1}{c}{JSON-Grinder} &
\multicolumn{1}{c}{Set-Based} & \multicolumn{1}{c}{LSTM-Based} \\
\hline
automobile &  79.0\% (1.95) &\multicolumn{1}{c}{---}&  82.4\% (4.97) &  84.9\% (3.58)\\
bank &        79.3\% (0.44) &  63.2\% (15.75) &  78.9\% (0.60) &  79.3\% (0.42)\\
car &         79.9\% (1.40) &\multicolumn{1}{c}{---}& 100.0\% (0.00) & 100.0\% (0.00)\\
contraceptive & 55.3\% (2.78) &  55.5\% (5.17) &  54.0\% (5.03) &  53.4\% (2.31)\\
mushroom &   100.0\% (0.00) & 100.0\% (0.00) &  99.7\% (0.37) & 100.0\% (0.00)\\
nursery &     95.6\% (0.69) &\multicolumn{1}{c}{---}&  99.2\% (0.89) & 100.0\% (0.02)\\
seismic &     73.3\% (5.28) &\multicolumn{1}{c}{---}&  72.9\% (3.37) &  71.6\% (3.86)\\
student &     37.0\% (5.49) &  36.0\% (4.31) &  30.2\% (3.59) &  34.2\% (5.12)\\
\multicolumn{4}{c}{} \\
& \multicolumn{1}{c}{Linear} & \multicolumn{1}{c}{SVM} & \multicolumn{1}{c}{Random-Forest} &  \\
\hline
automobile &  72.7\% (4.20) &  77.6\% (3.90) &  78.5\% (6.43) & \\
bank &  79.3\% (0.67) &  80.1\% (0.56) &  79.0\% (0.52) &\\
car &  77.3\% (3.57) &  79.2\% (2.23) &  72.5\% (1.85) & \\
contraceptive &  51.3\% (2.35) &  55.2\% (3.25) &  49.1\% (2.44) & \\
mushroom & 100.0\% (0.00) & 100.0\% (0.00) & 100.0\% (0.00) & \\
nursery &  91.1\% (0.34) &  96.6\% (0.21) &  93.4\% (0.52) & \\
seismic &  74.5\% (4.38) &  72.5\% (2.97) &  70.2\% (5.80) & \\
student &  35.1\% (4.52) &  34.7\% (2.66) &  33.6\% (2.59) & \\

    \end{tabular}

    \caption{Average test accuracy across the four deep models and nine datasets, with standard deviation over five runs in brackets. }
    \label{tab:full_json_uci_results}
\end{table}

\newpage

\section{Hyper-parameters for experiments}

\begin{table}[!htbp]

    {\centering
    \setlength\tabcolsep{3.5pt}
    \begin{tabular}{r | c c c c | c c c c | c c c c | c c c c | c c c c }
    & \multicolumn{15}{c}{Fold} \\
    & \multicolumn{4}{c}{1} & \multicolumn{4}{c}{2} & \multicolumn{4}{c}{3} & \multicolumn{4}{c}{4} & \multicolumn{4}{c}{5} \\
    & E\footnotemark[1] &  BS\footnotemark[2] &  W\footnotemark[3] & L \footnotemark[4] &
    E &  BS &  W & L &   E &  BS &  W & L &   E &  BS &  W & L &   E &  BS &  W & L \\
    \hline
     automobile &   30 &  16 & 128 &   3 &   30 &  16 & 128 &   1 &   30 &   4 & 128 &   1 &   50 &   4 &  64 &   1 &   30 &  16 & 128 &   3 \\
           bank &   3 &  64 & 128 &   3 &   4 &  16 &  64 &   1 &   1 &   4 &  64 &   3 &   4 &   4 &  32 &   5 &   4 &   4 &  64 &   3 \\
            car &   10 &   4 &  32 &   3 &   3 &   4 & 128 &   1 &   5 &  16 &  64 &   1 &   10 &   4 &  32 &   3 &   20 &   4 & 128 &   5 \\
  contraceptive &   3 &   4 &  32 &   5 &   3 &   4 &  64 &   3 &   3 &   4 & 128 &   5 &   30 &  64 &  32 &   3 &   10 &  64 &  64 &   5 \\
       mushroom &   1 &   4 &  32 &   1 &   1 &   4 &  32 &   1 &   1 &   4 &  32 &   1 &   1 &   4 &  32 &   1 &   2 &   4 &  32 &   1 \\
        nursery &   4 &  16 &  64 &   5 &   4 &  64 & 128 &   3 &   4 &   4 &  64 &   3 &   3 &  16 & 128 &   3 &   4 &   4 &  64 &   1 \\
        seismic &   5 &  16 &  64 &   5 &   10 &  64 &  64 &   5 &   5 &  64 &  32 &   1 &   3 &  16 &  64 &   1 &   10 &  64 &  64 &   3 \\
        student &   5 &   4 &  32 &   1 &   20 &  16 &  32 &   1 &   20 &   4 & 128 &   5 &   3 &   4 &  64 &   1 &   10 &  64 & 128 &   5 \\

    \end{tabular}}

    \footnotemark[1]{Number of epochs} \\
    \footnotemark[2]{Batch size} \\
    \footnotemark[3]{Hidden layers width} \\
    \footnotemark[4]{Number of layers}

    \caption{Chosen hyper-parameters for the multi-layer perceptron baseline}
    \label{tab:json_uci_parameters_mlp}
\end{table}

%bags
%TODO
\begin{table}[!htbp]
    \centering
    \parbox{.7\textwidth}{\centering
    \begin{tabular}{r | c c | c c | c c | c c | c c  }
                & \multicolumn{10}{c}{Fold} \\
                & \multicolumn{2}{c}{1} & \multicolumn{2}{c}{2} & \multicolumn{2}{c}{3}
                & \multicolumn{2}{c}{4} & \multicolumn{2}{c}{5} \\
                &   BS\footnotemark[1] &   TS\footnotemark[2] &   BS &  TS &  BS &   TS &   BS &  TS &   BS &  TS \\
    \hline
           bank &    4 & 4000 &   4 & 9000 &  64 &  700 &  16 & 2000 &  16 & 2000 \\
  contraceptive &   16 &  500 &   4 & 6000 &  64 &  500 &  16 & 1500 &  16 & 2500 \\
       mushroom &   64 &  500 &  64 &  500 &  64 &  500 &  64 &  500 &  64 &  500 \\
        student &   64 &  100 &  64 &  100 &   4 & 1000 &   4 & 1000 &  64 &  100 \\
    \end{tabular}
    \begin{flushleft}
\footnotemark[1] - Batch size \\
\footnotemark[2] - Train iterations
    \end{flushleft}}

    \caption{Chosen hyper-parameters for the JSON-Grinder baseline}
    \label{tab:json_uci_parameters_bags}
    %\vspace{-2cm}
\end{table}

\begin{table}[!htbp]
    %\vspace{-2cm}
    {\centering
    \begin{tabular}{r | c c c | c c c | c c c | c c c | c c c  }
     & \multicolumn{15}{c}{Fold} \\
                &         &   1 &     &         &   2 &     &         &   3 &     &         &   4 &     &         &   5 &     \\
    & E\footnotemark[1] &  BS\footnotemark[2] &  W\footnotemark[3] &
                                             E &  BS &  W &      E &  BS &  W &      E &  BS &  W &      E &  BS &  W \\
    \hline
     automobile &   50 &   4 &  64 &   50 &  16 & 128 &   30 &   4 & 128 &   50 &   4 &  64 &   50 &   4 &  64 \\
           bank &    4 &  64 & 128 &    4 &   4 &  64 &    4 &  64 &  64 &    4 &  16 & 128 &    3 &   4 & 128 \\
            car &   20 &   4 &  32 &   20 &   4 &  64 &   20 &   4 &  32 &   20 &   4 &  32 &   10 &   4 &  64 \\
  contraceptive &   34 &   4 &  64 &   34 &   4 &  32 &   20 &  16 &  64 &   30 &  64 &  64 &   34 &  64 &  32 \\
       mushroom &    2 &   4 &  32 &    3 &   4 &  32 &    3 &   4 &  32 &    3 &   4 &  32 &    7 &   4 &  32 \\
        nursery &    4 &   4 &  64 &    4 &   4 &  32 &    3 &   4 &  64 &    2 &   4 & 128 &    2 &   4 &  64 \\
        seismic &   20 &   4 &  32 &   30 &   4 & 128 &   20 &  64 & 128 &   20 &   4 &  64 &   50 &  16 & 128 \\
        student &   50 &  64 &  32 &   50 &  64 &  32 &   20 &  64 &  64 &   20 &  16 &  64 &    5 &  16 & 128 \\

    \end{tabular}}

    \footnotemark[1]{Number of epochs} \\
    \footnotemark[2]{Batch size} \\
    \footnotemark[3]{Hidden layers width}

    \caption{Chosen hyper-parameters for the set-based JSON-NN model}
    \label{tab:json_uci_parameters_set}
\end{table}

%LSTM
\begin{table}[!htbp]

    {\centering
    \begin{tabular}{r | c c c | c c c | c c c | c c c | c c c  }
    & \multicolumn{15}{c}{Fold} \\
                &         &   1 &     &         &   2 &     &         &   3 &     &         &   4 &     &         &   5 &     \\
    & E\footnotemark[1] &  BS\footnotemark[2] &  W\footnotemark[3] &
                                             E &  BS &  W &      E &  BS &  W &      E &  BS &  W &      E &  BS &  W \\
    \hline
     automobile &   30 &   4 & 128 &   10 &   4 & 128 &   30 &   4 & 128 &   20 &   4 & 128 &   50 &   4 & 128 \\
           bank &    4 &  64 & 128 &    3 &  16 &  64 &    4 &  16 &  32 &    2 &   4 & 128 &    4 &   4 &  64 \\
            car &   10 &   4 &  32 &   10 &   4 &  32 &   10 &   4 &  32 &   10 &   4 &  32 &    5 &   4 &  32 \\
  contraceptive &    2 &   4 & 128 &    5 &  64 &  64 &    3 &  64 &  32 &   20 &   4 &  32 &   34 &  64 & 128 \\
       mushroom &    1 &   4 &  32 &    1 &   4 &  32 &    1 &   4 &  32 &    2 &   4 &  32 &    1 &   4 &  32 \\
        nursery &    2 &   4 &  32 &    3 &   4 &  32 &    2 &   4 &  32 &    2 &   4 &  32 &    2 &   4 &  32 \\
        seismic &   10 &  64 &  32 &    5 &  64 &  64 &    3 &   4 &  64 &    1 &  16 & 128 &    5 &  16 &  32 \\
        student &    3 &  16 & 128 &    5 &  16 &  64 &   10 &  16 &  64 &   10 &  16 &  32 &    5 &  16 &  64 \\

    \end{tabular}}

    \footnotemark[1]{Number of epochs} \\
    \footnotemark[2]{Batch size} \\
    \footnotemark[3]{Hidden layers width}

    \caption{Chosen hyper-parameters for the LSTM-based JSON-NN model}
    \label{tab:json_uci_parameters_lstm}
    %\vspace{-2cm}
\end{table}

\begin{table}[!htbp]

    \centering
    \begin{tabular}{l c | c c c c }
    & & \multicolumn{4}{c}{Dataset fraction} \\
    Model & Parameter & 5\% & 20\% & 50\% & 100\%\\
    \hline
    \multirow{4}{0.2\textwidth}{Multi-Layer Perceptron}
    & Epochs      & 100 & 100 & 50 & 50 \\
    & Batch size  &   4 &  16 &  4 &  4 \\
    & Hidden size & 128 & 128 & 32 & 32 \\
    & Num layers  &   5 &   3 &  5 &  3 \\
    \hline
    \multirow{2}{0.2\textwidth}{JSON-Grinder}
    & Train Steps & 150000 & 100000 & 250000 & 250000 \\
    & Batch size  &   4 &  4 &  4 &  4 \\
    \hline
    \multirow{3}{0.2\textwidth}{Set-Based JSON model}
    & Epochs      & 100 & 100 &  50 & 50 \\
    & Batch size  &   4 &  64 &   4 &  4 \\
    & Hidden size &  32 & 128 & 128 & 64 \\
    \hline
    \multirow{3}{0.2\textwidth}{TreeLSTM-Based JSON model}
    & Epochs      & 100 & 100 &  50 & 50 \\
    & Batch size  &  64 &   4 &  16 &  4 \\
    & Hidden size & 128 & 128 & 128 & 32 \\
    \hline
    \multirow{3}{0.2\textwidth}{Tailored JSON model}
    & Epochs      &  50 &  50 &  50 & 30 \\
    & Batch size  &  4  &  16 &   4 &  64 \\
    & Hidden size &  64 & 128 &  64 & 128 \\
    \end{tabular}

    \caption{Chosen hyper-parameter for all models for the different fractions of the poker hands dataset. }
    \label{tab:json_uci_parameters_poker}
\end{table}

\newpage

\section{Examples of constructed JSON entries for UCI datasets}\label{ap:uci_examples}

\subsection{Car evaluation}
\begin{lstlisting}
{
  "PRICE": {
    "buying": "vhigh",
    "maint": "vhigh"
  },
  "TECH": {
    "COMFORT": {
      "doors": 2,
      "persons": 2,
      "lug_boot": "small"
    },
    "safety": "low"
  }
}
\end{lstlisting}

\subsection{Nursery evaluation}
\begin{lstlisting}
{
  "EMPLOY": {
    "parents": "usual",
    "has_nurs": "proper"
  },
  "STRUCT_FINAN": {
    "STRUCTURE": {
      "form": "complete",
      "children": 1
    },
    "housing": "convenient",
    "finance": "convenient",
  },
  "SOC_HEALTH": {
    "social": "nonprob",
    "health": "recommended"
  }
}
\end{lstlisting}

\subsection{Poker hands}
\begin{lstlisting}
[
    {
        "suit": "Hearts",
        "rank": 10
    },
    {
        "suit": "Hearts",
        "rank": "Jack"
    },
    {
        "suit": "Hearts",
        "rank": "King"
    },
    {
        "suit": "Hearts",
        "rank": "Queen"
    },
    {
        "suit": "Hearts",
        "rank": "Ace"
    }
]
\end{lstlisting}

\subsection{Mushroom}
\begin{lstlisting}
{
    "cap": {
        "shape": "convex",
        "surface": "smooth",
        "color": "brown"
    },
    "gill": {
        "attachment": "free",
        "spacing": "close",
        "size": "narrow",
        "color": "black"
    },
    "stalk": {
        "shape": "enlarging",
        "root": "equal",
        "surface": {
            "above-ring": "smooth",
            "below-ring": "smooth"
        },
        "color": {
            "above-ring": "white",
            "below-ring": "white"
        }
    },
    "veil": {
        "type": "partial",
        "color": "white"
    },
    "ring": {
        "type": "pendant",
        "number": 1
    },
    "bruising": true,
    "odor": "pungent",
    "spore-print-color": "black",
    "population": "scattered",
    "habitat": "urban"
}
\end{lstlisting}

\subsection{Seismic bumps}
\begin{lstlisting}

{
    "work-shift": "N",
    "assessments": [
        {
            "type": "seismic",
            "result": "a"
        },
        {
            "type": "acoustic",
            "result": "a"
        },
        {
            "type": "geophone",
            "readings": {
                "total-energy": 15180,
                "deviation-energy": -72,
                "number-pulses": 48,
                "deviation-pulses": -72
            },
            "result": "a"
        }
    ],
    "readings": {
        "total-energy": 0,
        "max-energy": 0,
        "bumps": [
            {
                "range-start": 10e2,
                "range-end": 10e3,
                "total-bumps": 0
            },
            {
                "range-start": 10e3,
                "range-end": 10e4,
                "total-bumps": 0
            },
            {
                "range-start": 10e4,
                "range-end": 10e5,
                "total-bumps": 0
            },
            {
                "range-start": 10e5,
                "range-end": 10e6,
                "total-bumps": 0
            },
            {
                "range-start": 10e6,
                "range-end": 10e7,
                "total-bumps": 0
            },
            {
                "range-start": 10e7,
                "range-end": 10e8,
                "total-bumps": 0
            },
            {
                "range-start": 10e8,
                "range-end": 10e10,
                "total-bumps": 0
            }
        ],
        "total-bumps": 0
    }
}
\end{lstlisting}

\subsection{Contraceptive method choice}
\begin{lstlisting}
{
    "wife": {
        "age": 24,
        "education": 2,
        "religion-is-islam": true,
        "now-working": true
    },
    "husband": {
        "education": 3,
        "occupation": 2
    },
    "children": 3,
    "standard-of-living": 3,
    "media-exposure": 0
}
\end{lstlisting}

\subsubsection{Automobile}
\begin{lstlisting}
{
    "make": "alfa-romero",
    "price": 13495,
    "curb-weight": 2548,
    "mpg": {
        "city": 21,
        "highway": 27
    },
    "powertrain": {
        "engine": {
            "fuel-type": "gas",
            "fuel-system": "mpfi",
            "aspiration": "std",
            "engine-type": "dohc",
            "compression-ratio": 9.00,
            "bore": 3.47,
            "stroke": 2.68,
            "num-of-cylinders": 4,
            "displacement": 130,
            "horsepower": 111,
            "peak-rpm": 5000
        },
        "engine-location": "front",
        "drive-wheels": "rwd"
    },
    "chassis": {
        "dimensions": {
            "length": 168.80,
            "width": 64.10,
            "height": 48.80
        },
        "wheel-base": 88.60,
        "body-style": "convertible",
        "num-of-doors": 2
    }
}
\end{lstlisting}

\subsection{Bank marketing}
\begin{lstlisting}
{
    "age": 56,
    "job": "housemaid",
    "marital-status": "married",
    "education": "basic.4y",
    "loan": {
        "personal": "no",
        "mortgage": "no",
        "in-default": "no"
    },
    "contact": {
        "type": "telephone",
        "last-contact": {
            "month": "may",
            "weekday": "mon"
        },
        "this-campaign": {
            "number": 1
        }
    },
    "indicators": {
        "emp.var.rate": 1.1,
        "cons.price.idx": 93.994,
        "cons.conf.idx": -36.4,
        "euribor3m": 4.857,
        "nr.employed": 5191
    }
}
\end{lstlisting}

\subsection{Student performance}
\begin{lstlisting}
{
    "school": "Gabriel Pereira",
    "reason-for-chosing": "course preference",
    "sex": "Female",
    "age": 18,
    "health": 3,

    "household": {
        "rural": false,
        "travel-time": "15 to 30 min.",
        "internet": false,
        "education-support": false,
        "family": {
            "size": "> 3",
            "relationship quality": 4,
            "parents": {
                "separated": true,
                "guardian": "mother",
                "mother": { "education": "higher education", "job": "stay-at-home" },
                "father": { "education": "higher education", "job": "teacher" }
            }
        }
    },
    "study": {
        "hours-per-week": "2 to 5 hours",
        "continue-to-higher": true,
        "attended-nursery": true,
        "extra-support": true,
        "num-fails": 0,
        "tutored": false,
        "absences": 4
    },
    "social": {
        "free-time": 3,
        "socialising-external": 4,
        "alcohol-consumption": { "weekday": 1, "weekend": 1 },
        "extra-curricular": false,
        "in-relationship": false
    }
}
\end{lstlisting}

%
%In our main JSON implementation (Secion~\ref{sec:json_implementation_main}) we also apply an additional linear transformation to $\mathbf h^{SumLSTM}$ to get our final $\mathbf h^{out}$:
%\begin{equation*}
%    \mathbf h^{out} = Lin(\mathbf h^{SumLSTM}, \theta^T_p)
%\end{equation*}
%

\end{appendices}

\end{document}